%% file: main.tex
\documentclass[preprint,5p,times,twocolumn]{elsarticle}

\usepackage{times}
\usepackage[numbers]{natbib}
\usepackage{multicol}
\usepackage[bookmarks=true]{hyperref}
\usepackage{graphicx} 
\usepackage{array}
\usepackage{xcolor}
\usepackage{float}
\usepackage{fancyhdr}
\usepackage{lipsum}
\usepackage{comment}
\usepackage{makecell}
\usepackage{pgfgantt}
\usepackage[utf8]{inputenc}
\usepackage{array}
\usepackage{colortbl}
\usepackage{xcolor}
\usepackage{url}
\usepackage{booktabs}
\usepackage{amsmath}
\usepackage{lipsum}
\usepackage{comment}
\usepackage{makecell}
\usepackage{pgfgantt}
\usepackage{amsmath} 
\usepackage{xcolor} 
\usepackage{booktabs}
\usepackage{adjustbox}
\usepackage{algpseudocode}
\usepackage{adjustbox}
\usepackage[ruled,vlined]{algorithm2e}
\usepackage{listings}
\usepackage{booktabs,multirow}
\usepackage[dvipsnames]{xcolor}
\usepackage{multirow}
\usepackage{array}
\usepackage{subcaption}
\usepackage{xcolor}
\usepackage{amsmath,amssymb}      

\newcommand{\inc}[1]{\textcolor{red}{\footnotesize(#1)\,$\uparrow$}}
\newcommand{\dec}[1]{\textcolor{ForestGreen}{\footnotesize(#1)\,$\downarrow$}}
\newcommand{\decl}[1]{\textcolor{red}{\footnotesize(#1)\,$\downarrow$}}
\newcommand{\nog}{\textcolor{gray}{\footnotesize (no impr.)}}
\newcommand{\incl}[1]{\textcolor{ForestGreen}{\footnotesize(#1)\,$\uparrow$}}

\algtext*{EndIf}
\algtext*{EndFor}
\algtext*{EndWhile}
\algtext*{EndProcedure}
\algtext*{EndFunction}

\journal{Information Fusion}

\begin{document}

\begin{frontmatter}



\title{Collaborative Trajectory Prediction via Late Fusion}

\author[a,d]{Nadya Abdel Madjid}
\author[a,d]{Murad Mebrahtu}
\author[e]{Zakhar Yagudin}
\author[b]{Bilal Hassan}
\author[a]{Naoufel Werghi}
\author[c,d]{Jorge Dias}
\author[e]{Dzmitry Tsetserukou}
\author[a,d]{Majid Khonji}

\affiliation[a]{organization={Khalifa University, Computer Science}, city={Abu Dhabi}, country={UAE}}
\affiliation[b]{organization={New York University Abu Dhabi, Computer Science}, city={Abu Dhabi}, country={UAE}}
\affiliation[c]{organization={Khalifa University, Computer and Information Engineering}, city={Abu Dhabi}, country={UAE}}
\affiliation[d]{organization={Khalifa University, KUCARS-KU Center for Autonomous Robotic Systems, Department of Computer Science}, city={Abu Dhabi}, country={UAE}}   
\affiliation[e]{organization={Skolkovo Institute of Science, Intelligent Space Robotics Laboratory, Center for Engineering Systems and Sciences, 
and Technology}, city={Moscow}, country={Russia}}

\begin{abstract}
Predicting future trajectories of surrounding traffic agents is critical for safe autonomous navigation and collision avoidance. Despite all advances in the trajectory forecasting realm, the prediction models remains vulnerable to uncertainty caused by occlusions, limited sensing range, and perception errors. Collaborative vehicle-to-vehicle (V2V) approaches help reduce this uncertainty by sharing complementary information. Existing collaborative trajectory prediction methods typically fuse feature maps at the perception stage to construct a holistic scene view. Further this holistic representation is decoded into the future trajectories. Such design incurs substantial communication overhead due to the exchange of high-dimensional feature representations and often assumes idealized bandwidth and synchronization, limiting practical deployment. We address these limitations by shifting collaboration from perception to the prediction module and introducing a late-fusion framework for shared forecasts. The framework is model-agnostic and treats collaborating vehicles as independent asynchronous agents. We evaluate the approach on the OPV2V, V2V4Real, and DeepAccident datasets, comparing individual and collaborative forecasting. Across all datasets, late fusion consistently reduces miss rate and improves trajectory success rate ($\mathrm{TSR}_{0.5}$), defined as the fraction of ground-truth agents with final displacement error below 0.5\,m. On the real-world V2V4Real dataset, collaborative prediction improves the success rate by $1.69\%$ and $1.22\%$ for both intelligent vehicles, respectively, compared with individual forecasting.
\end{abstract}







\end{frontmatter}


\input{sections/1_intro}
\input{sections/2_related_work}
\input{sections/3_methodology}
\input{sections/4_implementation}
\input{sections/5_results}
\input{sections/6_conclusion}




\bibliographystyle{elsarticle-num-names} 
\bibliography{references}





\end{document}

%% file: sections/1_intro.tex
\section{Introduction}
\label{sec:intro}

\begin{figure}[t!]
\centering
   \includegraphics[width=0.48\textwidth]{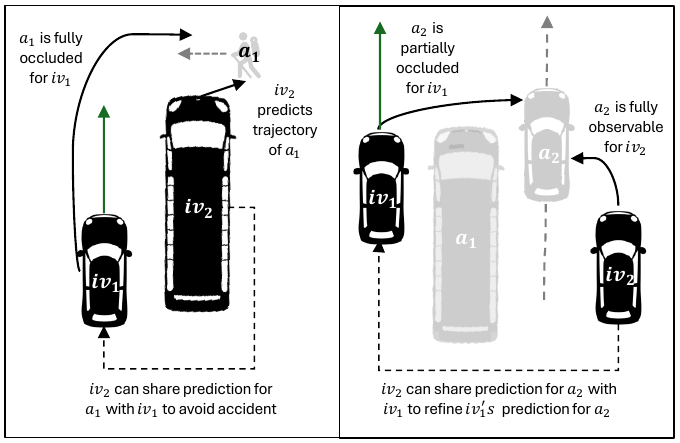}
\caption{Occlusion scenarios, where collaboration between vehicles is useful. Left scenario: a fully occluded agent is not observable to the ego vehicle (e.g., a pedestrian hidden behind a truck), motivating the need for external information to avoid collision. Right scenario: agent $a_2$ is partially occluded for $iv_1$ by agent $a_1$, degrading tracking and prediction, while $iv_2$ has a clear view and can estimate $a_2$'s future trajectory more reliably.}
   \label{fig:scenarios}
\end{figure}

Predicting future trajectories of surrounding traffic agents is essential for safe navigation and collision avoidance in autonomous driving. While single-vehicle trajectory forecasting has matured significantly \citep{zhao2020tnt, li2024planning, seokha2024}, it remains sensitive to perception errors caused by occlusions, sensor noise, and limited sensing range (some scenarios are illustrated in Figure \ref{fig:scenarios}). Collaboration between communicating vehicles, through sharing complementary information, can address these challenges. By fusing observations from different perspectives, vehicles can reason about agents that are otherwise unobservable or only partially observed, helping to reduce uncertainty in their predictions. However, enabling effective collaboration in a multi-vehicle system requires carefully balancing improvements in predictive accuracy against communication bandwidth, latency, and potential degradation caused by inaccurate or misaligned information fusion.

Most existing works \citep{10.1007/978-3-030-58536-5_36, d9, Xu2022CoBEVTCB} explore collaboration by fusing intermediate feature representations at the perception stage, aiming to construct a holistic view of the environment. This shared representation is then passed to downstream task modules, including predictor. In such approach, collaboration is effectively confined to the feature-fusion stage of perception, limiting exploration of integrating shared information directly into task-specific modules. Few existing trajectory forecasting models perform fusion outside the perception stage. V2X-Graph \citep{ruan2023learning} performs intermediate fusion over past trajectories using interpretable motion and interaction features. CMP model \citep{wu2024cmp} adopts a hybrid fusion strategy: it fuses BEV features during perception to enhance object detection and applies late fusion to refine predicted trajectories.

While it is intuitive that trajectory prediction benefits from access to a holistic scene view, in such pipelines the prediction module operates on a fixed fused scene representation and behaves as a single-vehicle forecaster with richer contextual input. We tackle this gap by introducing a dedicated fusion layer for explicit trajectory-level fusion across vehicles. Moving collaboration from perception-stage feature fusion to prediction-stage trajectory fusion aligns it closer with the forecasting objective and improves communication efficiency, since messages consist of sparse trajectory predictions rather than high-dimensional feature maps.

Next, most works \cite{lei2024robustcollaborativeperceptionexternal, Li2021LearningDC, xia2025plentypolymorphicfeatureinterpreter} do not model delay under the assumption that collaborating vehicles operate as independent distributed systems. Temporal misalignment arises from two sources. First, sensing, inference, and broadcasting occur asynchronously across vehicles, so shared information is already temporally behind the ego vehicle’s current inference time. Second, transmission latency introduces additional staleness. Many existing designs instead assume near-synchronized operation, with matched frame rates and negligible delay, so that shared features can be treated as if they align with the ego vehicle’s timestamp before inference. Such assumptions are infeasible in practical deployments, where system heterogeneity and communication latency are the norm.

Few SOTA models \citep{xu2025codyntrustrobustasynchronouscollaborative, 10.1007/978-3-031-19824-3_19} explicitly compensate for temporal misalignment. For example, \citet{yu2023flowbased} use a learned feature-flow module to extrapolate infrastructure features to the ego vehicle's current timestamp, while \citet{10.1007/978-3-031-19824-3_19} proposed SyncNet, which estimates delayed collaborative features from historical information.

In our design, each vehicle operates as an independent autonomous system with its own parameters and algorithmic stack. A dedicated collaboration layer handles asynchronous message arrivals and performs data association, ensuring that vehicles refer to the same physical agents and that shared forecasts are aligned to the ego vehicle’s prediction timeline. This temporal alignment block is algorithmic and does not require joint training to learn delay compensation.

Another challenge in collaborative setting is keeping communication requirements compatible with realistic bandwidth constraints. A number of works \citep{9197364, 10378239, 10.1145/3581783.3611699, yang2023howcomm, Hu2022Where2commCC, Chen_2023_ICCV} focus on optimizing communication efficiency aimig for message size reduction. For example, the CMP model \citep{wu2024cmp} reports a reduction to 0.23 MB/s, which under its assumption of a 100 ms synchronized processing window corresponds roughly to a 10 Hz broadcast rate and about 23 KB per message. More broadly in V2V settings, V2X-ViT \citep{10.1007/978-3-031-19842-7_7} reports approximately 120 KB per message, while TransIFF reports a compact communication footprint of $2^{12}$ bytes (4096 bytes). However, these studies do not clearly specify which communication protocol is assumed to reliably sustain such transmission rates in real-time operation. In contrast, CoSDH \citep{xu2025cosdhcommunicationefficientcollaborativeperception} explicitly references an IEEE 802.11p–based DSRC protocol with a reported transmission rate of 27 Mbps under point-to-point communication. Under these assumptions, the authors estimate an upper bound of 86 KB per message at 10 Hz when four vehicles collaborate. However, as the number of collaborating vehicles increases, the allowable message size decreases proportionally, raising concerns about the scalability. 

Another line of work focuses on query-driven collaboration. In such setup, the ego vehicle first identifies regions of uncertainty and requests complementary information from neighboring vehicles \cite{gao2025stamp, 9636761, Hong_2024_CVPR}. It broadcasts its query and waits for relevant responses before inference. Following this approach, Direct-CP \citep{tao2025directcpdirectedcollaborativeperception} introduces a direction-aware collaborative perception framework, where ego vehicle proactively generates query maps based on directional interest. While this strategy improves efficiency and implies transmission of information only for the most relevant regions, it needs inference-time analysis to assess the practical cost of awaiting responses. 

In our work, a late fusion approach is adopted as a scalable and practical strategy for collaborative trajectory forecasting. In late fusion, the amount of shared information, compared to early \citep{chen2023co} and intermediate fusion \citep{wei2023asynchronyrobust, 10030450, 9676458, yang2023spatiotemporal}, is reduced, lowering communication bandwidth requirements. By leveraging shared trajectories, late fusion refines predictions for partially occluded agents and enables reasoning about agents that are fully occluded and not locally observable. From a communication perspective, we adopt a broadcast–listen paradigm and explicitly specify V2V communication protocols that impose practical constraints on broadcasting frequency (10 Hz) and message size (up to 1,500 bytes). Vehicles do not establish fixed connections or explicitly allocate bandwidth. Messages are broadcast within communication range, and intelligent vehicles within the range acting as aggregators, listen and fuse the received predictions. 

Next, we detail the system design and the assumptions regarding communication topologies and constraints, summarize the main contributions of this study, and outline the structure of the paper.

\subsection{System Design and Assumptions}
Each intelligent vehicle is equipped with a full perception–prediction stack and operates on its own local clock. Each vehicle is assumed to have access to GPS time as a coarse global clock, providing a shared reference for aligning timestamps across vehicles without requiring matched frame rates. For coordinate representation, vehicles store and transmit data in a global coordinate system.

\textbf{\textit{Communication Protocol:}}
In the proposed collaborative trajectory prediction system, vehicles communicate using a broadcast-listen paradigm without establishing persistent point-to-point connections. This design avoids delays associated with connection setup and ensures compatibility with highly dynamic traffic scenarios, where vehicles may collaborate for both short and long durations.

C-V2X Mode~4, developed under the 3GPP standards (Release~14)\footnote{\url{https://www.3gpp.org/specifications-technologies/releases/release-14}}, enables direct vehicle-to-vehicle (V2V) communication based on 4G LTE technology. It supports single-hop broadcast communication, where vehicles transmit messages directly to others within range. Messages can be customized to include trajectory prediction--related information. C-V2X Mode~4 operates in the 5.9~GHz band with a typical channel bandwidth of 10~MHz and a broadcast frequency of 10~Hz (every 100~ms). The message size is typically on the order of 200--400~bytes, with a communication range of approximately 300--500~m in urban environments, around 800~m in suburban settings, and up to 1--2~km in rural areas.

Following the same design principles, i.e., single-hop direct communication and customizable message formats, 5G NR-V2X, developed under the 3GPP standards (Release~16)\footnote{\url{https://www.3gpp.org/specifications-technologies/releases/release-16}}, can also be deployed. 5G NR-V2X extends these capabilities using 5G technology and operates primarily in the 5.9~GHz band with a typical channel bandwidth of 10~MHz in the Sub-6~GHz spectrum. It supports broadcast communication at 10~Hz (every 100~ms), with message sizes typically ranging from 200 to 1,500~bytes and a communication range comparable to that of C-V2X Mode~4.

Accordingly, in our system design the upper bound supported by 5G NR-V2X is taken as the limit, assuming a maximum message size of 1,500~bytes at a broadcast frequency of 10~Hz to maintain low-latency communication.

\textbf{\textit{Broadcasting and Aggregating Modes:}}
Intelligent vehicles can operate in broadcasting and aggregating modes. In the broadcasting mode, the vehicles share their individual forecasts, while in aggregating mode, the vehicles collect and fuse data broadcast by other vehicles to refine their own predictions. The intelligent vehicles can act simultaneously as broadcasters and aggregators; however, in our single-hop setting we restrict communication to broadcasting individual, unfused predictions. Sharing trajectories after fusion would implicitly propagate aggregated information, since fused predictions already include information received from other vehicles. The impact of relaying and multi-hop communication is beyond the scope of this study and is left for future investigation.

\subsection{Contributions}
We claim the following contributions:
\begin{itemize}
    \item A collaborative trajectory prediction framework with a dedicated collaboration layer that models collaborating vehicles as independent systems in a distributed setting, together with a scalable late-fusion mechanism.
    
    \item A modular and reproducible implementation of the proposed framework, facilitating further research on collaborative trajectory prediction and alternative fusion strategies. The code is publicly available at \href{https://github.com/AV-Lab/Collab_Late_Trajectory_Prediction}{repository}.
\end{itemize}

\subsection{Paper Structure}
The rest of the paper is organized as follows. Section~\ref{sec:rel_work} reviews existing work in the area of single and collaborative trajectory forecasting. Section~\ref{sec:methodology} introduces the proposed collaborative trajectory prediction framework, covering the problem formulation, collaboration model, and fusion mechanism. Implementation details, the training setup, and the multi-vehicle datasets along with the evaluation protocol used in the experiments are presented in Section~\ref{sec:implementation}. Section~\ref{sec:results} reports the results of all experiments conducted to evaluate the effectiveness of the framework. Finally, Section~\ref{sec:conclusion} concludes the paper with a discussion and outlines future directions.

%% file: sections/2_related_work.tex
\section{Related Work}
\label{sec:rel_work}

The literature on collaborative vehicle systems for trajectory prediction remains sparse. Existing works investigate how information from multiple viewpoints can be aggregated to improve object detection \citep{chen2023co, Li2021LearningDC}, tracking \citep{9676458}, trajectory prediction \citep{10.1007/978-3-030-58536-5_36}, planning \citep{Cui2022CoopernautED, glaser2023communicationcritical}, or task-invariant \citep{li2022multirobot}. In this related work, we review existing collaborative models for trajectory prediction, as well as collaborative perception methods that address communication efficiency and transmission delays.

\subsection{Trajectory Prediction}

BEV-based cooperative prediction methods \citep{10.1007/978-3-030-58536-5_36, d9} rely on extracting BEV representations independently at each agent and then sharing intermediate feature maps for fusion. The pioneering model V2VNet \citep{10.1007/978-3-030-58536-5_36} following this approach, uses a cross-vehicle GNN to aggregate delay-compensated and pose-aligned features before decoding the fused representation into detection and trajectory forecasting outputs. V2XFormer \citep{d9} adopts a similar pipeline. It generates BEV features with a Swin-Transformer-based backbone, spatially aligns them to the ego frame, and decodes the fused BEV representation into forecasts. 

Another line of work \citep{zhang2025comtpcooperativetrajectoryprediction, ruan2023learning} aims to improve trajectory forecasting by compensating for incomplete past observations. Co-MTP \citep{zhang2025comtpcooperativetrajectoryprediction} performs vehicle--infrastructure cooperative fusion over both past trajectories and predicted motions. The model combines AV and infrastructure trajectories to enrich incomplete observations. It further integrates the ego planning trajectory with infrastructure-predicted trajectories of surrounding agents to model planning-oriented interactions. V2X-Graph \citep{ruan2023learning} performs forecasting-oriented fusion over historical trajectories. The problem of associating past trajectories belonging to the same agent across views is formulated as a trainable graph-link prediction task supervised by pseudo labels. After association, the model fuses motion and interaction representations before decoding multimodal forecasts.

Lastly, CMP \citep{wu2024cmp} combines fusion at both the perception and prediction stages to improve predicted trajectories. It performs intermediate BEV feature fusion for cooperative perception and late prediction aggregation across vehicles. Unlike earlier BEV-based approaches, which encode multiple past sweeps into a fused representation and decode prediction downstream, CMP is closer to our design: it first decodes detections from the fused perception output, then uses a tracker to recover past motion, and finally aggregates shared predictions across vehicles. The prediction model is pre-trained, and the aggregation module is subsequently trained with predictors' finetuning. Spatial alignment in its' cooperative perception part is handled by a learnable transformation operator.

\subsection{Communication Efficiency} 
Mitigating communication redundancy in collaborative perception has been explored in two main directions: (1) determining which information should be shared by surrounding agents to complement the ego vehicle’s observations \citep{9197364, 10378239, 10.1145/3581783.3611699, yang2023howcomm, xu2025cosdhcommunicationefficientcollaborativeperception}, and (2) designing efficient compression mechanisms to minimize message sizes while preserving essential information \citep{Hu2022Where2commCC, Chen_2023_ICCV, hu2024communicationefficientcollaborativeperceptioninformation, ding2025point}. The first line of research typically follows a request--response pipeline: the ego agent broadcasts a query, surrounding agents evaluate their local information with respect to that request, and only selected complementary responses are transmitted back for fusion. The second line of research focuses on compressing communicated representations while retaining task-relevant context. Exsisting approaches include sparse transmission of high-confidence features \citet{Hu2022Where2commCC}, filtering low-value predictions before transmission \citet{Chen_2023_ICCV}, codebook-based message compression \citet{hu2024communicationefficientcollaborativeperceptioninformation}, and compact object-level representations that preserve semantic and structural information \citet{ding2025point}.

\subsection{Transmission Delays} 
Existing methods compensate for stale collaborative information through predictive temporal alignment at the feature level, aiming to realign delayed features to the ego vehicle's current timestamp before fusion. Among detection methods, SyncNet \citep{10.1007/978-3-031-19824-3_19} jointly estimates current collaborative features and collaboration attention from historical information and uses latency-conditioned modulation for synchronization. FFNet \citep{yu2023flowbased} uses a learned feature-flow representation to predict infrastructure features at the target time. CoDynTrust \citep{xu2025codyntrustrobustasynchronouscollaborative} combines uncertainty-aware ROI evaluation with motion-based feature warping to suppress unreliable asynchronous features during alignment.

In collaborative trajectory prediction models, V2VNet \citep{10.1007/978-3-030-58536-5_36} uses a trainable delay-compensation module, CMP \citep{wu2024cmp} handles delay through synchronized 100 ms processing and one-frame-delayed prediction aggregation, and V2X-Graph \citep{ruan2023learning} relies on interpolation-based synchronization of delayed past trajectories.

%% file: sections/3_methodology.tex
\section{Methodology}
\label{sec:methodology}

\begin{figure*}[t!]
\centering
   \includegraphics[width=1.0\textwidth]{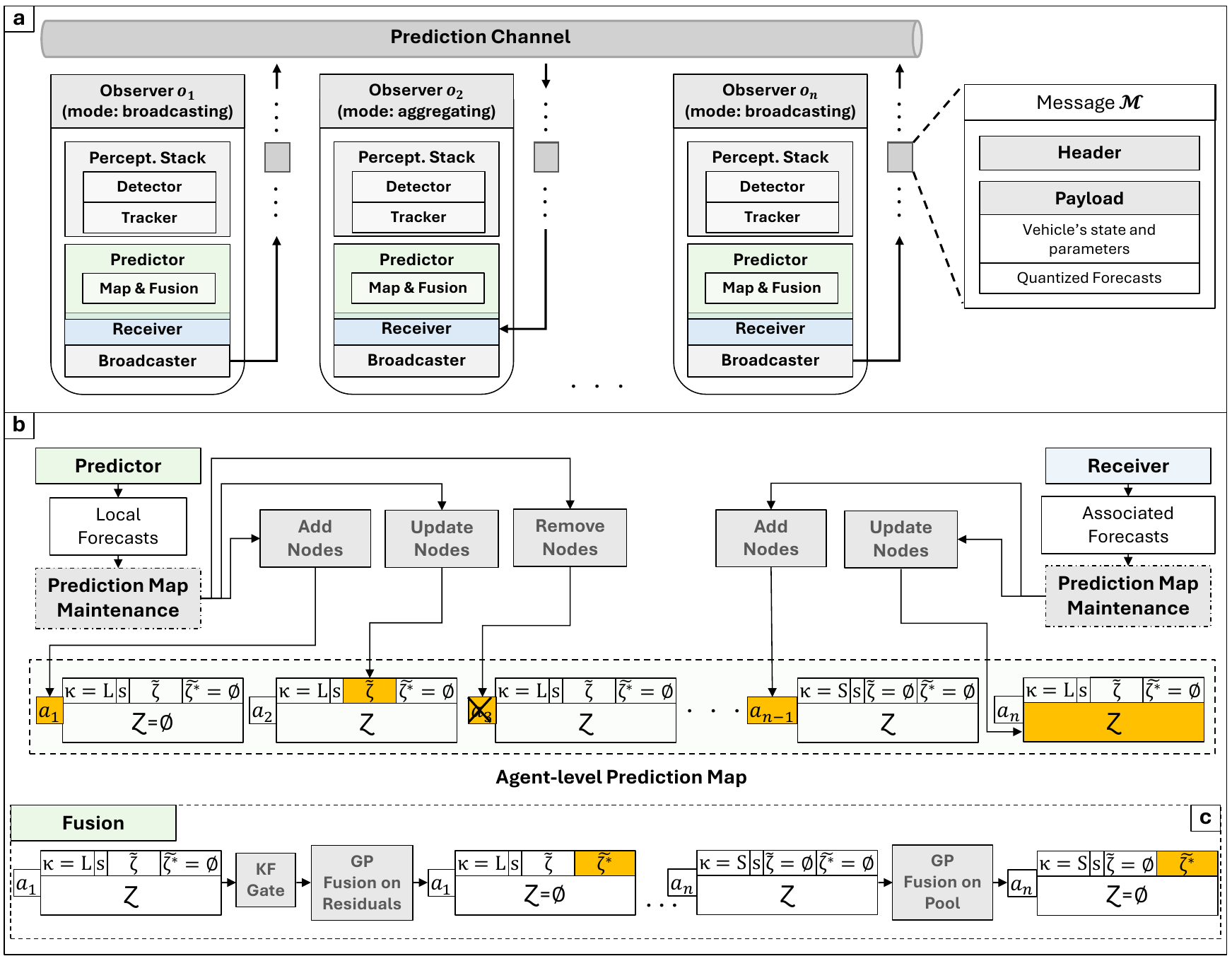}
   \caption{System architecture of the proposed collaborative trajectory prediction framework. (a) Broadcast--listen communication setup and packet structure used to share predicted trajectories between vehicles. (b) Agent-level prediction map update: the perception--prediction stack adds, updates, and removes nodes; after data association, the prediction pools of Category~L (locally observed) nodes are updated, while Category~S (shared-only) nodes are added or updated. The fusion module combines trajectories for Category~L and Category~S nodes and resets their prediction pools after each fusion step.}
   \label{fig:implementation_sketch}
\end{figure*}

This section presents the problem formulation of collaborative trajectory prediction, including the collaboration model and objective, and describes the fusion mechanism (refer to Figure \ref{fig:implementation_sketch} for overview).

\subsection{Problem Formulation}

Let $\mathcal{A}$ be a set of agents, among which $\mathcal{C} \subseteq \mathcal{A}$ is the set of collaborative agents (e.g., intelligent vehicles), and $\mathcal{N} = \mathcal{A} \setminus \mathcal{C}$ is the set of non-collaborative agents (e.g., pedestrians, cyclists, human-driven vehicles). For brevity, we refer to non-collaborative agents as targets and to the collaborative agents observing them as observers. Each agent $a \in \mathcal{A}$ has a state $s_a(t) \in \mathbb{R}^d$ at time $t$. A typical state representation includes the bird's-eye-view position, and optionally, heading angle, and longitudinal velocity. Each observer $i \in \mathcal{C}$ is configured with an observation horizon $T^{\mathrm{obs}}_i$ and a prediction horizon $T^{\mathrm{pred}}_i$. At any given time $t$, a target $k \in \mathcal{N}$ may be visible by a subset of observers. The observed trajectory of target $k$ by observer $i$ is denoted as
\[
  \tilde{\tau}_{ki}
  =
  \big(
    \tilde{s}_{ki}(t - T^{\mathrm{obs}}_{i} + 1),\,
    \tilde{s}_{ki}(t - T^{\mathrm{obs}}_{i} + 2),\,
    \dots,\,
    \tilde{s}_{ki}(t)
  \big).
\]

Based on the observed trajectory $\tilde{\tau}_{ki}$, observer $i$ predicts the future trajectory of target $k$ denoted
\[
  \tilde{\zeta}_{ki}
  =
  \big(
    \tilde{s}_{ki}(t+1),\,
    \tilde{s}_{ki}(t+2),\,
    \dots,\,
    \tilde{s}_{ki}(t + T^{\mathrm{pred}}_{i})
  \big),
\]
where $\tilde{s}_{ki}(\cdot)$ denotes predicted states at future time steps. The corresponding true future trajectory of target $k$ is
\[
  \zeta_k
  =
  \big(
    s_k(t+1),\,
    s_k(t+2),\,
    \dots,\,
    s_k(t + T^{\mathrm{pred}}_i)
  \big).
\]

In a collaborative setting, different observers that see the same target $k$ will each produce their own prediction, so that multiple predicted trajectories are available for the same underlying true trajectory $\zeta_k$. The goal of collaborative fusion is to combine these observer-specific predictions into a refined estimate of $\zeta_k$. We assume that an observer $i$ can provide an estimate of its prediction uncertainty for target $k$, encoded as a variance of the predicted trajectory per sample, (i.e., each trajectory is presented by ($\tilde{\zeta}_{ki}$, $\tilde{\Sigma}_{ki}$)), derived from some uncertainty estimation techniques. When we refer to $\tilde{\zeta}$, we assume that $\tilde{\Sigma}$ is always available.

\subsubsection{Collaboration Model}

We consider a collaboration model where observers share their predicted trajectories and associated uncertainty estimates over a V2V broadcast channel. Each observer $i \in \mathcal{C}$ maintains a local agent-level prediction map $\Pi_i(V_i)$, where $V_i \subseteq \mathcal{A}$ denotes the set of targets known to $i$, including targets it directly observes and targets it learns about through information received from other observers. For a target $k$, the corresponding entry in the map is represented by an attribute tuple
\begin{equation}
  \Pi_i(k) = \bigl(\kappa_{ki}, s_{ki}(t), \tilde{\zeta}_{ki}, \tilde{\zeta}^{*}_{ki}, \mathcal{Z}_{ki}\bigr),
\end{equation}
where $\kappa_{ki}$ is a node category, $s_{ki}(t)$ is the current state estimate of target $k$ at time $t$ as maintained by observer $i$, $\tilde{\zeta}_{ki}$ is the local prediction, $\tilde{\zeta}^{*}_{ki}$ is the fused trajectory, and $\mathcal{Z}_{ki}$ denotes the prediction pool. The pool $\mathcal{Z}_{ki}$ collects all remote predictions for target $k$ aggregated from other observers by observer $i$. 

Entries in $\Pi_i$ can belong to two categories:

\paragraph{\textbf{Category L} (local targets)}
An entry with $\kappa_{ki} = L$ corresponds to a target that is directly observed by the observer $i$ (i.e., through its perception-prediction stack). For such targets:
\begin{enumerate}
  \item The local state estimate $s_{ki}(\cdot)$ and prediction $\tilde{\zeta}_{ki}$ is maintained ans stored in $\Pi_i(k)$;
  \item The fused trajectory $\tilde{\zeta}^{*}_{ki}$ is obtained by applying the fusion to $\tilde{\zeta}_{ki}$ and the elements of $\mathcal{Z}_{ki}$.
\end{enumerate}

\paragraph{\textbf{Category S} (shared-only targets)}
An entry with $\kappa_{ki} = S$ corresponds to a target that is not currently observed by observer $i$ (e.g., due to occlusion or perception failure) and is known only through predictions received from other observers. For such targets:
\begin{enumerate}
  \item The local prediction field in $\Pi_i(k)$ is empty, i.e., $\zeta_{ki} = \emptyset$, and the current state $s_{ki}(t)$ is estimated from aligned remote predictions (e.g., state estimate $s_{kj}(\cdot)$ shared by another observer $j$);
  \item The fused trajectory $\tilde{\zeta}^{*}_{ki}$ is obtained by applying the fusion mechanism to the elements of $\mathcal{Z}_{ki}$;
  \item A target of \textit{Category~S} is changed to target of \textit{Category~L} if it becomes directly observable by observer $i$.
\end{enumerate}

For consistent fusion and updates of entries in the prediction map $\Pi_i$, the collaboration layer is equipped with a data association mechanism that performs temporal alignment and spatial matching of received trajectories. Temporal alignment is performed with respect to the current prediction timestamp of observer $i$. Upon receiving a trajectory $\tilde{\zeta}_{kj}$ shared by observer $j$ for target $k$, all trajectory samples that lie before observer $i$’s prediction timestamp are discarded. The current shared state $s_{kj}(\cdot)$ is set to the first remaining sample, and the timestamps of the rest samples are re-anchored so that the trajectory is expressed relative to observer $i$’s prediction timeline.

The temporally aligned states and trajectories broadcast by observer $j$ are then passed to a spatial matching step. After matching, the association module outputs a matched set of targets $M_i^j \subseteq \mathcal{N}$, whose predictions from $j$ are associated with existing entries in $\Pi_i$, and an unmatched set of targets $U_i^j \subseteq \mathcal{N}$, whose predictions from $j$ do not match any current entry in $\Pi_i$. These predictions are candidates for instantiating new shared-only entries with $\kappa=S$.

As a remark, the data association mechanism is independent of the fusion layer and can be replaced by more advanced or simplified variants. Its main role is to ensure that entries in $\Pi_i$ consistently refer to the same physical agents across different observers and that shared trajectories are temporally aligned.

Initially, $\Pi_i$ is empty. As observer $i$ begins to track targets locally and generate forecasts, entries in $\Pi_i$ are created and maintained. After receiving a broadcast message from observer $j \in \mathcal{C}$, and data association, observer $i$ merges the received predictions into its prediction map $\Pi_i$ (refer to Algorithms~\ref{alg:map_update_predictor} and~\ref{alg:map_update_association} for details).

For local targets according to the update rule:
\begin{equation}
  \mathcal{Z}_{pi} := \mathcal{Z}_{pi} \cup \big\{(\tilde{\zeta}_{pj}, \Sigma_{pj})\big\},
   p \in M_i^j,
  \label{eq:predpool_update}
\end{equation}
and for shared nodes according to the add rule:
\begin{equation}
  \Pi_i(p) := \bigl(\kappa_{pi}=S,\; s_{pi}(t)= s_{pj}(t),\; \mathcal{Z}_{pi}=\{(\tilde{\zeta}_{pj}, \Sigma_{pj})\}\bigr),
   p \in U_i^j.
  \label{eq:new_shared_entry}
\end{equation}

\begin{algorithm}[h!]
\caption{Prediction Map Update by Predictor}
\label{alg:map_update_predictor}
\DontPrintSemicolon
\KwIn{%
\begin{tabular}[t]{@{}l@{}}
prediction map $\Pi_i(V_i)$\\
local predictions $\mathcal{P}=\bigl((\text{id}_q,\; s_q(t),\; \tilde{\zeta}_q,\; \tilde{\Sigma}_q)\bigr)_{q=1}^{N}$\\
match threshold $\lambda_{\text{match}}$
\end{tabular}%
}
\KwOut{updated map $\Pi_i$}
\BlankLine
$V_L \gets \{ v \in V_i \mid \kappa(v) = L \}$ \\
$V_S \gets \{ v \in V_i \mid \kappa(v) = S \}$ \\
$\mathcal{P}_{\text{matched}} \gets \emptyset$\;
$\mathcal{P}_{\text{unmatched}} \gets \mathcal{P}$\;
\BlankLine
\ForEach{$p \in \mathcal{P}$}{
  \If{$\exists v \in V_i : \text{id}(v) = p.\text{id}$}{
    $v.s \gets p.s(t)$\;
    $v.\tilde{\zeta} \gets (p.\tilde{\zeta},p.\tilde{\Sigma})$\;
    $\mathcal{P}_{\text{matched}} \gets \mathcal{P}_{\text{matched}} \cup \{p.\text{id}\}$\;
    $\mathcal{P}_{\text{unmatched}} \gets \mathcal{P}_{\text{unmatched}} \setminus p$\;
  }
}
\BlankLine
\ForEach{$p \in \mathcal{P}_{\text{unmatched}}$}{ 
  $v^\star \gets \arg\min_{v \in V_S} \text{distance}\!\bigl(p.s(t), v.s\bigr)$\;
  \If{$\text{distance}\!\bigl(p.s(t), v^\star.s\bigr) \le \lambda_{\text{match}}$}{
    $\kappa(v^\star) \gets L$\;
    $\text{id}(v^\star) \gets p.\text{id}$\;
    $v^\star.s \gets p.s(t)$\;
    $v^\star.\tilde{\zeta} \gets (p.\tilde{\zeta},p.\tilde{\Sigma})$\;
    $\mathcal{P}_{\text{matched}} \gets \mathcal{P}_{\text{matched}} \cup \{p.\text{id}\}$\;
    $\mathcal{P}_{\text{unmatched}} \gets \mathcal{P}_{\text{unmatched}} \setminus p$\;
  }
}
\BlankLine
\ForEach{$v \in V_i$}{
  \If{$\kappa(v) = L \wedge \text{id}(v) \notin \mathcal{P}_{\text{matched}} $}{
    remove $v$ from $\Pi_i$\;
  }
  \ElseIf{$\kappa(v) = S$ \text{and} $v$ \text{is unmatched for} $n$ \text{steps}}{
    remove $v$ from $\Pi_i$\;
  }
}
\BlankLine
\ForEach{$p \in \mathcal{P}_{\text{unmatched}}$}{
  create $v_{\text{new}}$ with $(p.id,\, s{=}p.s(t),\, \tilde{\zeta}{=}(p.\tilde{\zeta},p.\tilde{\Sigma}),\, \tilde{\zeta}^{*}{=}\emptyset,\, \kappa{=}L,\, \mathcal{Z}{=}\emptyset)$\;
  add $v_{\text{new}}$ to $\Pi_i$\;
}
\end{algorithm}

\begin{algorithm}[h!]
\caption{Prediction Map Update by Association}
\label{alg:map_update_association}
\DontPrintSemicolon
\KwIn{%
\begin{tabular}[t]{@{}l@{}}
prediction map $\Pi_i$\\
matched set $M=\{(\text{matched\_id}_q,\; s_q(t),\; \tilde{\zeta}_q,\; \tilde{\Sigma}_q)\}_{q=1}^{N_M}$\\
unmatched set $U=\{(s_r(t),\; \tilde{\zeta}_r,\; \tilde{\Sigma}_r)\}_{r=1}^{N_U}$\\
relevance threshold $\lambda_{\mathrm{rel}}$
\end{tabular}
}
\KwOut{updated map $\Pi_i$}
\BlankLine

\ForEach{$p\in M$}{
  $v \gets \Pi_i[p.\text{matched\_id}]$\;
  $v.\mathcal{Z} \gets v.\mathcal{Z} \cup \{(p.\tilde{\zeta},p.\tilde{\Sigma})\}$ 
}

\ForEach{$p \in U$}{
  \If{$\text{is\_relevant}\!\bigl(p.s(t)\bigr) > \lambda_{\mathrm{rel}}$}{
    create $v_{\text{new}}$ with
    $(\kappa{=}S,\; s{=}p.s(t),\; \tilde{\zeta}{=}\emptyset,\; \tilde{\zeta}^{*}{=}\emptyset,\; \mathcal{Z}{=}\{(p.\tilde{\zeta},p.\tilde{\Sigma})\})$
    add $v_{\text{new}}$ to $\Pi_i$\;
  }
}
\end{algorithm}

\subsubsection{Objective}
Given the prediction map $\Pi_i(k)$ for each target $k \in V_i$, the objective is to design a fusion scheme $\mathcal{F}$ that produces a refined trajectory prediction
\begin{equation}
  \tilde{\zeta}^{*}_{ki}
  =
  \mathcal{F}\big(\Pi_i(k)\big),
  \label{eq:fused_traj}
\end{equation}
such that $\tilde{\zeta}^{*}_{ki}$ is closer to the true future trajectory $\zeta_k$ than any individual prediction $\tilde{\zeta}_{kj} \in \Pi_i(k)$, as measured by a displacement error.

\subsection{Fusion Mechanism}

As fusion operator $\mathcal{F}$, a Gaussian Process (GP) regression framework~\cite{rasmussen2006gaussian} is adopted. GP regression explicitly models observation noise, which allows to incorporate the uncertainty estimates shared by observers alongside their predictions. Additionally, because a GP can be queried at any set of timestamps, it can fuse trajectories with different prediction horizons and sampling rates stored in $\Pi_i$ without requiring resampling to a common time grid.

Depending on the category type, two strategies are introduced: for \textit{Category L} nodes, fusion is performed through a GP-based residual correction of the ego trajectory, triggered by the KF-gate, and for \textit{Category S} nodes the fused trajectory is constructed directly from the pool of shared trajectories using GP. Since each trajectory sample $\tilde{s}_{kj}(t) \in \mathbb{R}^d$ is multi-dimensional (e.g., position, heading, velocity), one approach is to perform fusion with a multi-output GP over the full state. To keep inference lightweight, we instead apply independent scalar GP regression to each state component. In the common case where the state is a 2D position $s=(x,y)$, the fused trajectory is modeled as two independent scalar-valued functions of time, $f_x(t)$ and $f_y(t)$. If an observer provides a non-diagonal covariance over $(x,y)$, we use only the marginal variances (diagonal entries) for each coordinate and ignore cross-dimensional covariance terms.

Let $\mathcal{O}_i(k)$ denote the set of observers who observe target $k$ and whose predictions have been aggregated by observer $i$. Expanding Eq.~\ref{eq:fused_traj}:
\begin{equation}
\begin{aligned}
  \tilde{\zeta}^{*}_{ki}
  &=
  \mathcal{F}\Bigl(\mathcal{Z}_{ki} \cup \bigl\{(\tilde{\zeta}_{ki}, \tilde{\Sigma}_{ki})\bigr\}\Bigr) \\
  &=
  \mathcal{F}\Bigl(
    \bigl\{(\tilde{\zeta}_{k j}, \tilde{\Sigma}_{k j}) \,\big|\, j \in \mathcal{O}_i(k)\bigr\}
    \cup \bigl\{(\tilde{\zeta}_{ki}, \tilde{\Sigma}_{ki})\bigr\}
  \Bigr).
\end{aligned}
\label{eq:fused_traj_expand}
\end{equation}

The fusion is estimated independently in each coordinate:
\begin{equation}
\begin{aligned}
  \tilde{\zeta}^{*,x}_{ki}
  &=
  \mathcal{F}_x\Bigl(
    \bigl\{(\tilde{\zeta}^{x}_{k j}, \tilde{\Sigma}^{x}_{k j}) \,\big|\, j \in \mathcal{O}_i(k)\cup\{i\}\bigr\}
  \Bigr), \\
  \tilde{\zeta}^{*,y}_{ki}
  &=
  \mathcal{F}_y\Bigl(
    \bigl\{(\tilde{\zeta}^{y}_{k j}, \tilde{\Sigma}^{y}_{k j}) \,\big|\, j \in \mathcal{O}_i(k)\cup\{i\}\bigr\}
  \Bigr).
\end{aligned}
\label{eq:fused_traj_xy}
\end{equation}
where $\tilde{\zeta}^{x}_{kj}$ and $\tilde{\zeta}^{y}_{kj}$ denote the $x$- and $y$-coordinate sequences of $\tilde{\zeta}_{kj}$, and $\tilde{\Sigma}^{x}_{kj}$ and $\tilde{\Sigma}^{y}_{kj}$ denote the corresponding per-sample variances extracted from $\tilde{\Sigma}_{kj}$. 

For simplicity, we denote coordinate-wise sequences $\{\tilde{\zeta}^{x}_{k j}\}_{j \in \mathcal{O}_i(k)\cup\{i\}}$ and $\{\tilde{\zeta}^{y}_{k j}\}_{j \in \mathcal{O}_i(k)\cup\{i\}}$ as observation vectors
$X_{\mathrm{obs}} \in \mathbb{R}^N$ and $Y_{\mathrm{obs}} \in \mathbb{R}^N$. Let $T_{\mathrm{obs}}$ denote the timestamps associated with the entries of $X_{\mathrm{obs}}$ and $T_{\mathrm{inf}}$ denote the query timestamps at which observer $i$ requires the fused prediction (e.g., $i$'s local prediction timestamps).

Following the GP regression framework, the fused coordinates are modeled as random functions:
\begin{equation*}
\begin{aligned}
  f_x(t) &\sim \mathcal{GP}\!\bigl(m_x(t),\,k_{\theta_x}(t,t')\bigr), \\
  f_y(t) &\sim \mathcal{GP}\!\bigl(m_y(t),\,k_{\theta_y}(t,t')\bigr),
\end{aligned}
\label{eq:prior_xy}
\end{equation*}
where $t$ and $t'$ denote arbitrary time instants, $m_x(t)$ and $m_y(t)$ are mean functions, and $k_{\theta_x}(t,t')$ and $k_{\theta_y}(t,t')$ are temporal covariance kernels parameterized by hyperparameters $\theta_x$ and $\theta_y$, respectively.

The hyperparameters are re-estimated at each fusion step by maximizing the GP log marginal likelihood for each coordinate:
\[
\begin{aligned}
  \theta_x &= \arg\max_{\theta}\ \log p\!\left(X_{\mathrm{obs}} \mid T_{\mathrm{obs}}, \theta\right), \\
  \theta_y &= \arg\max_{\theta}\ \log p\!\left(Y_{\mathrm{obs}} \mid T_{\mathrm{obs}}, \theta\right).
\end{aligned}
\]

The associated uncertainties are incorporated as Gaussian observation noise via diagonal matrices, so that larger variances downweight uncertain samples in the GP posterior:
\begin{equation*}
\begin{aligned}
  \Sigma_{\mathrm{obs}}^{x} = \mathrm{diag}\!\Bigl(\bigl[\tilde{\Sigma}^{x}_{k j}\bigr]_{j \in \mathcal{O}_i(k)\cup\{i\}}\Bigr), \:\:
  \Sigma_{\mathrm{obs}}^{y} = \mathrm{diag}\!\Bigl(\bigl[\tilde{\Sigma}^{y}_{k j}\bigr]_{j \in \mathcal{O}_i(k)\cup\{i\}}\Bigr),
\end{aligned}
\label{eq:sigma_obs}
\end{equation*}
where $\bigl[\cdot\bigr]_{j\in \mathcal{O}_i(k)\cup\{i\}}$ denotes stacking the per-sample variance sequences in the same order as $X_{\mathrm{obs}}$. Through computing the posterior mean at $T_{\mathrm{inf}}$, the fused trajectories are obtained as:
\begin{equation}
\begin{aligned}
  \tilde{\zeta}^{*,x}_{ki}
  &=
  K\!\bigl(T_{\mathrm{inf}}, T_{\mathrm{obs}}\bigr)
  \Bigl[
    K\!\bigl(T_{\mathrm{obs}}, T_{\mathrm{obs}}\bigr)
    + \Sigma_{\mathrm{obs}}^{x}
  \Bigr]^{-1}
  X_{\mathrm{obs}}, \\
  \tilde{\zeta}^{*,y}_{ki}
  &=
  K\!\bigl(T_{\mathrm{inf}}, T_{\mathrm{obs}}\bigr)
  \Bigl[
    K\!\bigl(T_{\mathrm{obs}}, T_{\mathrm{obs}}\bigr)
    + \Sigma_{\mathrm{obs}}^{y}
  \Bigr]^{-1}
  Y_{\mathrm{obs}},
\end{aligned}
\label{eq:posterior_xy}
\end{equation}
where $K(\cdot,\cdot)$ denotes covariance matrices computed from the corresponding kernel ($k_{\theta_x}$ for $x$ and $k_{\theta_y}$ for $y$).

\subsubsection{Fusion for \textit{Category L} Nodes}

Let $u_{ki}(t)$ denote the trace of the Kalman filter covariance for target $k$ at time $t$, and let $u^{\mathrm{med}}_{ki}$ denote the running median of $u_{ki}(\cdot)$ over the tracklet history maintained by observer $i$. The covariance ratio at time $t$ is defined as
\[
  \rho_{\mathrm{cov}}(t) = \frac{u_{ki}(t)}{u^{\mathrm{med}}_{ki}}.
\]

Let $\ell_{ki}(t)$ denote the prediction-only streak, i.e., the number of consecutive KF steps up to time $t$ without a detection update. The KF-gate is defined as
\begin{equation*}
g_{ki}(t) =
\begin{cases}
1, & \text{if } \ell_{ki}(t) \ge \lambda_{\mathrm{min\_streak}}
\ \ \text{or}\ \ 
\rho_{\mathrm{cov}}(t) \ge \lambda_{\mathrm{cov\_ratio}}, \\[2pt]
0, & \text{otherwise},
\end{cases}
\label{eq:kf_gate}
\end{equation*}

where $\lambda_{\mathrm{min\_streak}}$ and $\lambda_{\mathrm{cov\_ratio}}$ are threshold hyperparameters.
The intuition is that when the KF runs consecutively in prediction-only mode without detection updates, the uncertainty increases because the state is propagated without new observations. If $g_{ki}(t)=0$, the fused output is set to the local prediction ($\tilde{\zeta}^{*}_{ki}\leftarrow \tilde{\zeta}_{ki}$), indicating that the vehicle is confident in its track and does not require fusion; otherwise, GP-based residual fusion is applied.

Once the KF-gate is open, fusion is performed as a residual correction. For each coordinate, the ego prediction is first interpolated at the timestamps of the pooled shared samples, and residual samples are formed by subtracting the interpolated ego prediction from the pooled predictions. The GP posterior mean is then computed using Eq.~\ref{eq:posterior_xy} with these residuals as observations, yielding an estimated correction over the ego query horizon. The fused trajectory is obtained by adding this correction to the ego prediction. The same procedure is applied independently in the $y$-dimension.

\subsubsection{Fusion for \textit{Category S} Nodes}

For these nodes, there is no local trajectory, so the fused trajectory is reconstructed entirely from the shared predictions stored in the pool~$\mathcal{Z}$. Fusion is triggered whenever at least one shared trajectory exists in the pool. Before fitting the GPs, the pooled $x$- and $y$-coordinate samples are normalized to zero mean and unit variance for numerical stability, and the associated per-sample variances are rescaled by the same normalization factor. The GP posterior mean is then computed using Eq.~\ref{eq:posterior_xy} at the ego vehicle’s prediction timestamps on the normalized values. The resulting fused trajectory is obtained by de-normalizing the posterior mean (i.e., applying the inverse scale-and-shift), independently for each coordinate.

%% file: sections/4_implementation.tex
\section{Implementation Details}
\label{sec:implementation}

Each collaborative vehicle is configurable with its own set of parameters (e.g., frame rate, forecasting frequency). The implemented stack follows a standard detect--track--predict paradigm: each module specifies the data format it consumes and produces, and wrapper layers convert between algorithm-specific representations and a common internal format. This design enables plug-and-play replacement of different detectors, trackers, and predictors without modifying the rest of the pipeline.

All collaborating vehicles are subscribed to a single prediction channel. In broadcasting mode, based on the configured broadcast frequency, collaborative vehicles construct a compact V2V message with quantized predictions. Messages are serialized using a lightweight binary format and sent via a ZeroMQ PUB/SUB socket, with optional compression to further reduce bandwidth. In aggregating mode, a vehicle fetches incoming messages, performs dequantization, data association, and triggers the prediction map update in parallel with the main perception--prediction stack.

To simulate collaboration, sensor data for all vehicles are pre-loaded and replayed using a shared simulation clock. At each tick, the current timestamp is provided to each vehicle, which runs detection, tracking, prediction, and communication according to its configured rates. Vehicles are advanced sequentially for reproducibility with incrementing the clock when full cycle is completed; extending the framework to parallel execution is left as future work.

Further, we detail the perception and predictor configurations, elaborate on the datasets used for evaluation, and describe the performance metrics.

\subsection{Perception Setup}
\label{sec:preception_setup}

The perception stack is implemented in two settings: (i) a controlled replay setting for simulated datasets, where occlusion scores are explicitly computed and used to filter precomputed detections, so that occlusions are reflected in the history passed to the predictor and their impact (and the benefit of collaboration) can be evaluated in isolation, and (ii) a real-deployment setting for V2V4Real dataset. The first setting reduces confounding effects from detector--tracker errors and focuses the analysis on occlusion-induced failures. Tracking and prediction are executed online within the stack, while detections are generated and stored offline; during runtime, they are loaded at each vehicle’s configured frame rate.

\subsubsection{Controlled Setup}
Fully occluded objects are not perceived by the ego vehicle. For heavily partially occluded objects, with a high probability the perception will miss it. A casting-ray algorithm is used to compute an occlusion score for each bounding box to approximate the behavior of a realistic detector. Occlusion scores are computed once during dataset preprocessing and stored per box. At runtime, when detections are loaded, all boxes with an occlusion score above $0.75$ are discarded before tracking and prediction (this threshold is chosen arbitrarily to represent cases where detection is expected to fail).

Let $\mathcal{B}$ denote the set of bounding boxes. For every target box $B_m \in \mathcal{B}$, its angular span $[\theta_L,\theta_R]$ is obtained by taking the ego-vehicle origin as a reference point, computing the viewing angle from the ego vehicle to each ground-plane corner of $B_m$, and using the smallest and largest of these angles as $\theta_L$ and $\theta_R$, respectively. After computing the span, $N$ rays with angles $\{\theta_n\}_{n=1}^N$ are cast, uniformly sampled in this interval. For each ray, the distance $d_m$ from $s_{\text{ego}}(t)$ to $B_m$ is compared with the distance $d_o$ to any other box $B_o \in \mathcal{B} \setminus {B_m}$ that intersects the same ray. A ray is marked as occluded if there exists an occluder $B_o$ that lies closer to the ego vehicle than the target box and is tall enough, relative to their distances, to cover the target from the ego vehicle’s viewpoint. Algorithm~\ref{alg:occlusion_score} shows the computation of the occlusion score $O_{L1}(B_m)$, defined as the fraction of rays that are occluded, where $h_m$ and $h_o$ denote the heights (vertical extents) of the boxes $B_m$ and $B_o$.

\begin{algorithm}[!h]
\caption{Casting-ray Algorithm}
\label{alg:occlusion_score}
\DontPrintSemicolon
\KwIn{$s_{ego}(t)$, $\mathcal{B}$, $N$}
\KwOut{$\{\,O_{L1}(B_m)\,\}_{B_m \in \mathcal{B}}$}
\BlankLine
\For{$B_m \in \mathcal{B}$}{
  compute $[\theta_L,\theta_R]$ of $B_m$ \;
  sample rays $\{\theta_n\}_{n=1}^N$ in $[\theta_L,\theta_R]$\;
  $c \gets 0$\;
  \For{$n = 1,\dots,N$}{
    compute $d_m = \lVert s_{\text{ego}}(t) - s_m(t) \rVert_2$ \;
    \For{$B_o \in \mathcal{B} \setminus \{B_m\}$}{
      \If{$\theta_n \cap B_o \neq \varnothing$}{
        compute $d_o = \lVert s_{\text{ego}}(t) - s_o(t) \rVert_2$ \;
        \If{$d_o < d_m$ and $h_o \geq h_m \cdot (d_o / d_m)$}{
          $c \gets c + 1$\;
          \textbf{break}\;
        }
      }
    }
  }
  $O_{L1}(B_m) \gets c / N$\;
}
\Return{$\{\,O_{L1}(B_m)\,\}_{B_m \in \mathcal{B}}$}\;
\end{algorithm}

To isolate association errors and focus on the drift introduced purely by the tracker’s state estimation, acompined tracker uses ground-truth IDs for association, bypassing appearance or geometry matching and directly linking detections across frames via provided object identifiers. For state estimation and prediction, each object is represented by a lightweight 2D Kalman filter with a 6D constant-acceleration state $\mathbf{x} = [x,y,v_x,v_y,a_x,a_y]^\top$ and measurements $\mathbf{z} = [x,y]^\top$, where $x$ and $y$ denote the bird’s-eye-view (ground-plane) position, $v_x$ and $v_y$ denote the corresponding velocity components, and $a_x$ and $a_y$ denote the corresponding acceleration components.

The tracker uses standard tracklet management: detections are assigned to existing tracks with the same GT ID and used in a predict--update step, or they initialize a new track when no matching ID exists. Tracks without an associated detection perform predict-only updates and are removed once a miss counter exceeds a fixed lifetime. Each tracklet stores a fixed-length history and updates the box dimensions $(d_x,d_y,d_z)$ using an exponential moving average. In addition, each tracklet maintains a streak counter $\ell$  to count consecutive predict-only steps and the running median covariance trace $u$, which are used for fusion gating.

\subsubsection{Real-Deployment Setup}
For 3D object detection, a LiDAR-based CenterPoint detector with a PointPillars backbone is used to detect vehicles, pedestrians, and cyclists. The point cloud is pillarized with a voxel resolution of $0.24 \times 0.24 \times 6.0$~m over a perception range of $[-60, 60]$~m. The model is initialized from a Waymo-pretrained checkpoint and fine-tuned on V2V4Real for 50 epochs with batch size 24 using single-sweep LiDAR input. During inference, non-maximum suppression with IoU threshold $0.2$ and a confidence threshold of $0.5$ are applied. Detections are generated offline on an NVIDIA RTX 3090 (24~GB) and replayed per frame during fusion.

The real-deployment tracker uses the same 6-state KF formulation, augmented with global data association, track lifecycle management, and improved initialization. For association, Euclidean distances between detections and active tracklets are computed as a distance gate. In parallel, an NIS-based cost matrix is formed for all track--detection pairs, where NIS is the squared Mahalanobis distance between the KF-predicted position and the detection, and the Hungarian algorithm is applied to obtain one-to-one assignments. Each candidate match is required to have a consistent class label and to pass two gates: a category- and track-state-dependent Euclidean distance threshold with different limits for tentative and confirmed tracks, and a fixed NIS threshold. Tracks are confirmed after a minimum number of hits and pruned after a fixed number of consecutive misses, while the initial velocity is optionally bootstrapped from the nearest same-class detection in the previous frame.

\subsection{Training Predictor} 
\label{sec:predictor_train}

To equip each predicted trajectory with an uncertainty estimate $\Sigma$, the predictor models future motion as a sequence of Gaussian-distributed displacements with diagonal covariance. At the current time $t$, the predictor outputs the mean and covariance of the displacement distribution for each future prediction timestamp $t' \in \{t+1,\dots,t+T^{\mathrm{pred}}\}$,
\[
\mu(t') = [\hat{d}_x(t'),\, \hat{d}_y(t')], 
\qquad
\Sigma(t') = 
\begin{bmatrix}
\sigma_{x}^2(t') & 0 \\
0 & \sigma_{y}^2(t')
\end{bmatrix}.
\]

To estimate predictive uncertainty, the model is trained using Gaussian negative log-likelihood loss \citep{10.5555/3295222.3295387}:
\[
\mathcal{L}_{\mathrm{NLL}}
=
\frac{1}{T^{\mathrm{pred}}}
\sum_{t'=t+1}^{t+T^{\mathrm{pred}}}
\Biggl[
\frac{\bigl(d_x(t') - \hat{d}_x(t')\bigr)^2}{2\sigma_{x}^2(t')}
+
\frac{\bigl(d_y(t') - \hat{d}_y(t')\bigr)^2}{2\sigma_{y}^2(t')} +
\Biggr.
\]
\[
\Biggl.
+
\frac{\log \sigma_{x}^2(t') + \log \sigma_{y}^2(t')}{2}
\Biggr].
\]

The loss penalizes prediction error relative to the predicted uncertainty: if the model inflates the variances, the $\log \sigma^2$ terms dominate and increase the loss; if it makes the variances too small, the squared-error terms scaled by $1/\sigma^2$ become large and increase the loss. This trade-off discourages both overconfident and underconfident predictions and encourages calibrated uncertainty estimates. Adversarial training is applied by adding a gradient-sign perturbation to the past-motion input and optimizing a weighted sum of the clean and adversarial negative log-likelihood losses.

Absolute positions are reconstructed by integrating the predicted displacements over the prediction horizon starting from the position at time $t$.

\subsection{Datasets}

The proposed fusion approach is evaluated on three V2V datasets that support trajectory forecasting: DeepAccident \cite{d9}, OPV2V \cite{d3}, and V2V4Real \cite{d7}. The DeepAccident dataset, simulated in CARLA, focuses on diverse collision scenarios with multi-modal data from four vehicles. The OPV2V dataset has up to 7 collaborating vehicles per scenario. V2V4Real is a real-world dataset collected on urban and highway routes, with data recorded from two vehicles.

All datasets are preprocessed into a unified structure using scenario identifiers as the primary index and intelligent vehicle identifiers as the second-level keys. Each vehicle entry stores sensors data, calibration parameters, ego-state information and annotations. 

\subsection{Evaluation Metrics}
\label{sec:metrics}

At each timestamp $t$, $\mathcal{B}_g(t)$ denotes the set of ground-truth bounding boxes and
$\mathcal{B}_p(t)$ denotes the set of predicted bounding boxes for which forecasts are available.
Ground-truth and predicted boxes are associated using IoU-based greedy matching: for each
$B_m \in \mathcal{B}_g(t)$, the predicted box with the highest IoU is selected if the maximum IoU
exceeds a threshold $\lambda_{\mathrm{match}}$:
\[
B_n^\star = \arg\max_{B_n\in\mathcal{B}_p(t)}\ \mathrm{IoU}(B_m,B_n),
\]
\[
\mathrm{match}(B_m)=
\begin{cases}
B_n^\star, & \text{if }\mathrm{IoU}\!\bigl(B_m,B_n^\star\bigr)\ge \lambda_{\mathrm{match}},\\
\text{None}, & \text{otherwise.}
\end{cases}
\]

After matching at timestamp $t$, let $\mathcal{K}(t)$ denote the set of matched ground-truth targets.
Displacement errors are computed over $\mathcal{K}(t)$. For each $k \in \mathcal{K}(t)$, let
$\zeta_k = \bigl(s_k(t+1),\dots,s_k(t+T^{\mathrm{pred}})\bigr)$ denote the ground-truth future trajectory and
$\tilde{\zeta}_k^{*} = \bigl(\tilde{s}_k^{*}(t+1),\dots,\tilde{s}_k^{*}(t+T^{\mathrm{pred}})\bigr)$ denote the fused future trajectory.
The timestamp-level ADE and FDE are computed as:
\[
\mathrm{ADE}(t)
=
\frac{1}{|\mathcal{K}(t)|\,T^{\mathrm{pred}}}
\sum_{k \in \mathcal{K}(t)}
\sum_{t'=t+1}^{t+T^{\mathrm{pred}}}
\left\| \tilde{s}_k^{*}(t') - s_k(t') \right\|_2,
\]
\[
\mathrm{FDE}(t)
=
\frac{1}{|\mathcal{K}(t)|}
\sum_{k \in \mathcal{K}(t)}
\left\| \tilde{s}_k^{*}(t+T^{\mathrm{pred}}) - s_k(t+T^{\mathrm{pred}}) \right\|_2.
\]

The miss rate at timestamp $t$ is computed as the fraction of unmatched ground-truth objects:
\[
\mathrm{MR}(t)
= \frac{|\mathcal{B}_g(t)| - |\mathcal{K}(t)|}{|\mathcal{B}_g(t)|}.
\]

To jointly capture changes in both accuracy and coverage, the trajectory success rate (TSR) is reported. TSR counts a target as successful if it is matched and its final displacement is below a threshold $\lambda_{\mathrm{tsr}}$. It is normalized by the total number of ground-truth targets, so unmatched targets are treated as failures.

Let $\mathbf{1}[\cdot] \in \{0,1\}$ denote the indicator function, equal to $1$ if the condition holds and $0$ otherwise.
The TSR at timestamp $t$ is defined as
\[
\mathrm{TSR}(t)
=
\frac{1}{|\mathcal{B}_g(t)|}
\sum_{k \in \mathcal{K}(t)}
\mathbf{1}\!\Bigl[
\bigl\| \tilde{s}_k^{*}(t+T^{\mathrm{pred}}) - s_k(t+T^{\mathrm{pred}}) \bigr\|_2
\le \lambda_{\mathrm{tsr}}
\Bigr].
\]
In all experiments, $\lambda_{\mathrm{tsr}}$ is set to $0.5\,\mathrm{m}$, denoted as $\mathrm{TSR}_{0.5}$.

At every timestamp $t$, all metrics $\mathrm{ADE}(t)$, $\mathrm{FDE}(t)$, $\mathrm{MR}(t)$, and $\mathrm{TSR}(t)$ are computed and aggregated over all scenarios in the dataset. The reported numbers for each experiment are averages over timestamps.

%% file: sections/5_results.tex
\section{Results}
\label{sec:results}

\begin{table*}[!t]
\centering
\setlength{\tabcolsep}{5pt}
\renewcommand{\arraystretch}{1.2}

\resizebox{\textwidth}{!}{%
\begin{tabular}{
|>{\raggedright\arraybackslash}p{3.3cm}
|cccc
|cccc
|cccc
|cccc|
}
\hline
\multicolumn{1}{|c|}{}
& \multicolumn{8}{c|}{\textbf{Valid subset}}
& \multicolumn{8}{c|}{\textbf{Test subset}} \\
\hline

\multicolumn{1}{|c|}{}
& \multicolumn{4}{c|}{\textbf{Vehicle 1}}
& \multicolumn{4}{c|}{\textbf{Vehicle 2}}
& \multicolumn{4}{c|}{\textbf{Vehicle 1}}
& \multicolumn{4}{c|}{\textbf{Vehicle 2}} \\
\cline{2-17}


\multicolumn{1}{|c|}{}
& ADE$\downarrow$ & FDE$\downarrow$ & MR$\downarrow$ & TSR$_{0.5}\uparrow$ 
& ADE$\downarrow$ & FDE$\downarrow$ & MR$\downarrow$ & TSR$_{0.5}\uparrow$
& ADE$\downarrow$ & FDE$\downarrow$ & MR$\downarrow$ & TSR$_{0.5}\uparrow$ 
& ADE$\downarrow$ & FDE$\downarrow$ & MR$\downarrow$ & TSR$_{0.5}\uparrow$ \\
\hline

\makecell[l]{\textbf{No fusion} \\ \textbf{(w/o Occ.), \textit{tr12}}}         
  & 0.694 & 1.681 & -     & 55.17 
  & 0.812 & 1.926 & -     & 56.92 
  & 1.245   & 3.023   & -   & 34.36   
  & 1.175   & 2.862   & -   & 35.69  \\

\makecell[l]{\textbf{No fusion} \\ \textbf{(under Occ.), \textit{tr12}}}        
  & \makecell{0.773 \\ \inc{+0.079}} & \makecell{1.849 \\ \inc{+0.168}}  & 30.42 & \makecell{36.49 \\ \decl{-18.68\%}}  
  & \makecell{0.894 \\ \inc{+0.082}} & \makecell{2.062 \\ \inc{+0.136}} & 34.69 & \makecell{33.68 \\ \decl{-23.24\%}} 
  & \makecell{1.350 \\ \inc{+0.105}}  & \makecell{3.202 \\ \inc{+0.179}}  & 21.97   & \makecell{24.72  \\ \decl{-9.64\%}} 
  & \makecell{1.259 \\ \inc{+0.084}}  & \makecell{2.982 \\ \inc{+0.100}}   & 24.21   & \makecell{25.91 \\ \decl{-9.78\%}}  \\

\textbf{Fusion, \textit{tr12:tr12}}          
  & \makecell{0.770 \\ \dec{-0.003}} & \makecell{1.841 \\ \dec{-0.008}} & \makecell{27.62 \\ \dec{-2.8\%}}     & \makecell{38.22 \\ \incl{+1.73\%}}  
  & \makecell{0.882 \\ \dec{-0.012}} & \makecell{2.043 \\ \dec{-0.019}} & \makecell{31.39  \\ \dec{-3.3\%}}    & \makecell{35.74 \\ \incl{+2.06\%}}  
  & \makecell{1.343  \\ \dec{-0.007}}  & \makecell{3.189 \\ \dec{-0.013}}  & \makecell{18.85 \\ \dec{-3.12\%}}    & \makecell{25.12   \\ \incl{+0.4\%}}  
  & \makecell{1.271 \\ \inc{+0.012}}  & \makecell{3.004 \\ \inc{+0.022}}  & \makecell{20.73  \\ \dec{-3.48\%}}  & \makecell{26.99 \\ \incl{+1.08\%}}    \\

\textbf{Fusion, \textit{tr12:lstm12}}        
  & \makecell{0.763 \\ \dec{-0.010}} & \makecell{1.825 \\ \dec{-0.024}}  & \makecell{27.62 \\ \dec{-2.8\%}} & \makecell{38.25 \\ \incl{+1.76\%}}  
  & \makecell{0.882 \\ \dec{-0.012}} & \makecell{2.042 \\ \dec{-0.020}} & \makecell{31.39 \\ \dec{-3.3\%}} & \makecell{35.69 \\ \incl{+2.01\%}} 
  & \makecell{1.335 \\ \dec{-0.015}}  & \makecell{3.172 \\ \dec{-0.030}}  & \makecell{18.84 \\ \dec{-3.13\%}}   & \makecell{25.18  \\ \incl{+0.46\%}} 
  & \makecell{1.260 \\ \inc{+0.001}}  & \makecell{2.983 \\ \inc{+0.001}}   & \makecell{20.73 \\ \dec{-3.48\%}}   & \makecell{26.96 \\ \incl{+1.05\%}}   \\

\textbf{Fusion, \textit{tr12:tr23}}      
  & \makecell{0.776 \\ \inc{+0.003}} & \makecell{1.852 \\ \inc{+0.003}}  & \makecell{27.62 \\ \dec{-2.8\%}}  & \makecell{38.25 \\ \incl{+1.76\%}}   
  & \makecell{0.887 \\ \dec{-0.007}} & \makecell{2.050 \\ \dec{-0.012}} & \makecell{31.28 \\ \dec{-3.41\%}} & \makecell{35.83 \\ \incl{+2.15\%}} 
  & \makecell{1.346 \\ \dec{-0.004}}  & \makecell{3.191 \\ \dec{-0.011}}  & \makecell{18.82 \\ \dec{-3.15\%}}  & \makecell{25.11  \\ \incl{+0.39\%}} 
  & \makecell{1.269 \\ \inc{+0.010}}  & \makecell{2.996 \\ \inc{+0.014}}   & \makecell{20.68 \\ \dec{-3.53\%}}   & \makecell{26.95 \\ \incl{+1.04\%}}   \\ \hline

\makecell[l]{\textbf{No fusion} \\ \textbf{(w/o Occ.), \textit{tr23}}}         
  & 1.348 & 3.248 & -     & 48.50 
  & 1.544 & 3.694 & -     & 51.05 
  & 2.000   & 4.972   & -   & 30.65   
  & 1.841   & 4.572   & -   & 31.40  \\ 

\makecell[l]{\textbf{No fusion} \\ \textbf{(under Occ.), \textit{tr23}}}         
  & \makecell{1.530 \\ \inc{+0.182}} & \makecell{3.640 \\ \inc{+0.392}} & 29.04  & \makecell{32.56 \\ \decl{-15.94\%}} 
  & \makecell{1.804 \\ \inc{+0.260}} & \makecell{4.206 \\ \inc{+0.512}} & 33.28 & \makecell{30.32 \\ \decl{-20.73\%}} 
  & \makecell{2.369 \\ \inc{+0.369}}  & \makecell{5.725 \\ \inc{+0.753}}   & 18.83   & \makecell{22.88 \\ \decl{-7.77\%}}   
  & \makecell{2.116 \\ \inc{+0.275}}  & \makecell{5.123 \\ \inc{+0.551}}   & 22.99   & \makecell{23.09 \\ \decl{-8.31\%}} \\ 

\textbf{Fusion, \textit{tr23:tr12}}      
  & \makecell{1.517 \\ \dec{-0.013}} & \makecell{3.658 \\ \inc{+0.018}}  & \makecell{26.57 \\ \dec{-2.47\%}} & \makecell{34.16 \\ \incl{+1.6\%}}  
  & \makecell{1.798 \\ \dec{-0.006}} & \makecell{4.275 \\ \inc{+0.069}} & \makecell{30.18 \\ \dec{-3.1\%}} & \makecell{32.14 \\ \incl{+1.82\%}} 
  & \makecell{2.364 \\ \dec{-0.005}}  & \makecell{5.795 \\ \inc{+0.070}}  & \makecell{16.78 \\ \dec{-2.05\%}}   & \makecell{23.11  \\ \incl{+0.23\%}} 
  & \makecell{2.145 \\ \inc{+0.029}}  & \makecell{5.266 \\ \inc{+0.143}}   & \makecell{19.67 \\ \dec{-3.32\%}}   & \makecell{23.96 \\ \incl{+0.9\%}}   \\
  
\textbf{Fusion, \textit{tr23:tr23}}          
  & \makecell{1.514 \\ \dec{-0.017}} & \makecell{3.607 \\ \dec{-0.033}} & \makecell{26.49 \\ \dec{-2.55\%}} & \makecell{34.13 \\ \incl{+1.57\%}} 
  & \makecell{1.783 \\ \dec{-0.021}} & \makecell{4.169  \\ \dec{-0.037}} & \makecell{30.07  \\ \dec{-3.21\%}} & \makecell{32.21  \\ \incl{+1.89\%}} 
  & \makecell{2.352 \\ \dec{-0.017}}   & \makecell{5.707 \\ \dec{-0.018}}  & \makecell{16.76   \\ \dec{-2.07\%}} & \makecell{23.15   \\ \incl{+0.27\%}}  
  & \makecell{2.118  \\ \inc{+0.002}}  & \makecell{5.129 \\ \dec{-0.006}}  & \makecell{19.63  \\ \dec{-3.36\%}}   & \makecell{23.95   \\ \incl{+0.86\%}}  \\

\hline

\makecell[l]{\textbf{No fusion} \\ \textbf{(w/o Occ.), \textit{lstm12}}}        
  & 0.688 & 1.606 & -     & 56.23 
  & 0.826 & 1.900 & -     & 57.09 
  & 0.905 & 2.132   & -   & 37.25 
  & 0.913   & 2.162   & -   & 37.48    \\

\makecell[l]{\textbf{No fusion} \\ \textbf{(under Occ.), \textit{lstm12}}}            
& \makecell{0.711 \\ \inc{+0.023}} & \makecell{1.672 \\ \inc{+0.066}}  & 30.42  & \makecell{36.42 \\ \decl{-19.81\%}} 
& \makecell{0.858 \\ \inc{+0.032}} & \makecell{1.957 \\ \inc{+0.057}} & 34.69 & \makecell{33.97 \\ \decl{-23.12\%}}
& \makecell{0.931 \\ \inc{+0.026}}  & \makecell{2.170 \\ \inc{+0.038}} & 21.07 & \makecell{27.18 \\ \decl{-10.07\%}} 
& \makecell{0.889 \\ \dec{-0.024}}  & \makecell{2.089 \\ \dec{-0.073}}  & 24.21  & \makecell{27.71  \\ \decl{-9.77\%}}  \\

  \textbf{Fusion, \textit{lstm12:lstm12}}          
  & \makecell{0.708 \\ \dec{-0.003}} & \makecell{1.667 \\ \dec{-0.005}} & \makecell{27.62 \\ \dec{-2.8\%}} & \makecell{38.18 \\ \incl{+1.76\%}}  
  & \makecell{0.852 \\ \dec{-0.006}} & \makecell{1.952 \\ \dec{-0.005}} & \makecell{31.41 \\ \dec{-3.28\%}} & \makecell{35.94 \\ \incl{+1.97\%}}  
  & \makecell{0.925 \\ \dec{-0.006}}  & \makecell{2.164 \\ \dec{-0.006}}  & \makecell{18.87 \\ \dec{-2.2\%}}  & \makecell{27.66   \\ \incl{+0.48\%}}  
  & \makecell{0.899 \\ \nog}  & \makecell{2.118 \\ \inc{+0.029}}   & \makecell{20.73  \\ \dec{-3.48\%}}  & \makecell{28.77  \\ \incl{+1.06\%}}   \\

\textbf{Fusion, \textit{lstm12:tr12}}        
  & \makecell{0.714 \\ \inc{+0.003}} & \makecell{1.684 \\ \inc{+0.012}} & \makecell{27.65 \\  \dec{-2.77\%}} & \makecell{38.15 \\ \incl{+1.73\%}}  
  & \makecell{0.853 \\ \dec{-0.005}} & \makecell{1.953 \\ \dec{-0.004}} & \makecell{31.39 \\ \dec{-3.3\%}} & \makecell{36.02 \\ \incl{+2.05\%}}  
  & \makecell{0.933 \\ \inc{+0.002}}  & \makecell{2.181 \\ \inc{+0.011}}  & \makecell{18.87\\ \dec{-2.2\%}}  & \makecell{27.60   \\ \incl{+0.42\%}}  
  & \makecell{0.910 \\ \inc{+0.011}}  & \makecell{2.138 \\ \inc{+0.049}}  & \makecell{20.71  \\ \dec{-3.5\%}} & \makecell{28.79 \\ \incl{+1.08\%}}   \\

  \textbf{Fusion, \textit{lstm12:lstm23}}      
  & \makecell{0.712 \\ \inc{+0.001}}  & \makecell{1.667 \\ \dec{-0.005}} & \makecell{27.65 \\  \dec{-2.77\%}} & \makecell{38.29 \\ \incl{+1.87\%}}  
  &\makecell{0.854 \\ \dec{-0.004}} & \makecell{1.942 \\ \dec{-0.015}}  & \makecell{31.30 \\ \dec{-3.39\%}} & \makecell{35.93 \\ \incl{+1.96\%}} 
  & \makecell{0.925 \\ \dec{-0.006}} & \makecell{2.156 \\ \dec{-0.014}} & \makecell{18.84 \\ \dec{-2.23\%}} & \makecell{27.58 \\ \incl{+0.4\%}} 
  & \makecell{0.898 \\ \inc{+0.009}}  & \makecell{2.101 \\ \inc{+0.012}}  & \makecell{20.69 \\ \dec{-3.52\%}}   & \makecell{28.80 \\ \incl{+1.09\%}}  \\ \hline

  \makecell[l]{\textbf{No fusion} \\ \textbf{(w/o Occ.), \textit{lstm23}}}        
  & 1.120 & 2.652 & -     & 10.10 
  & 1.275 & 2.973 & -     & 9.76 
  & 1.505   & 3.582   & -   & 16.23   
  & 1.480   & 3.557   & -   & 15.83  \\

  \makecell[l]{\textbf{No fusion} \\ \textbf{(under Occ.), \textit{lstm23}}}        
  & \makecell{1.181 \\ \inc{+0.061}} & \makecell{2.773 \\ \inc{+0.121}}  & 29.04  & \makecell{6.01 \\ \decl{-4.09\%}}  
  & \makecell{1.347 \\ \inc{+0.072}} & \makecell{3.116 \\ \inc{+0.143}} & 33.28 & \makecell{6.94 \\ \decl{-2.82\%}} 
  & \makecell{1.590 \\ \inc{+0.085}}  & \makecell{3.747 \\ \inc{+0.165}}  & 18.83   & \makecell{11.75  \\ \decl{-4.48\%}} 
  & \makecell{1.513 \\ \inc{+0.033}}  & \makecell{3.633 \\ \inc{+0.076}}   & 22.99   & \makecell{11.80 \\ \decl{-4.03\%}}   \\ 

\textbf{Fusion, \textit{lstm23:lstm12}}      
  & \makecell{1.195 \\ \inc{+0.014}} & \makecell{2.872 \\ \inc{+0.099}} & \makecell{26.57 \\ \dec{-2.47\%}}    & \makecell{7.77 \\ \incl{+1.76\%}} 
  & \makecell{1.365 \\ \inc{+0.018}}  & \makecell{3.246 \\ \inc{+0.130}}   & \makecell{30.21 \\ \dec{-3.07\%}}   & \makecell{8.77 \\ \incl{+1.84\%}}  
  & \makecell{1.607 \\ \inc{+0.017}}  & \makecell{3.885 \\ \inc{+0.138}} & \makecell{16.85 \\ \dec{-1.98\%}} & \makecell{12.12 \\ \incl{+0.37\%}} 
  & \makecell{1.562 \\ \inc{+0.049}}  & \makecell{3.846 \\ \inc{+0.213}}   & \makecell{19.62 \\ \dec{-3.37\%}} & \makecell{12.78 \\ \incl{+0.98\%}} \\
  
  \textbf{Fusion, \textit{lstm23:lstm23}}          
  & \makecell{1.176 \\ \dec{-0.005}} & \makecell{2.764 \\ \dec{-0.009}} & \makecell{26.57 \\ \dec{-2.47\%}} & \makecell{7.60 \\ \incl{+1.59\%}} 
  & \makecell{1.336 \\ \dec{-0.011}} & \makecell{3.094 \\ \dec{-0.022}} & \makecell{30.11 \\ \dec{-3.17\%}} & \makecell{8.63 \\ \incl{+1.69\%}} 
  & \makecell{1.580 \\ \dec{-0.010}}  & \makecell{3.737 \\ \dec{-0.010}}  & \makecell{16.82 \\ \dec{-2.01\%}}  & \makecell{12.05  \\ \incl{+0.3\%}}
  & \makecell{1.522 \\ \inc{+0.009}}  & \makecell{3.651 \\ \inc{+0.018}}  & \makecell{19.58 \\ \dec{-3.41\%}}   & \makecell{12.59 \\ \incl{+0.79\%}}  \\ 

\hline

\end{tabular}}
\caption{
OPV2V results for \textit{Vehicle 1} and \textit{Vehicle 2} on the validation and test subsets. Notation: \textit{tr12}/\textit{lstm12} denote Transformer/LSTM predictors using 1\,s history and a 2\,s prediction horizon; \textit{tr23}/\textit{lstm23} use 2\,s history and a 3\,s horizon. The tracker maintains tracklets for 10 frames in the \textit{12} setting and 20 frames in the \textit{23} setting. For \textit{No fusion (under Occ.)}, the second line shows the change relative to \textit{No fusion (w/o Occ.)}. For \textit{Fusion} rows, the second line shows the change relative to \textit{No fusion (under Occ.)} for the same ego predictor. ADE/FDE deltas are absolute, MR and TSR$_{0.5}$ are relative percentage changes (\%).
}
\label{tab:opv2v_table}
\end{table*}

\begin{table*}[t]
\centering
\setlength{\tabcolsep}{3.8pt}
\renewcommand{\arraystretch}{1.12}

\resizebox{\textwidth}{!}{%
\begin{tabular}{
|c|>{\raggedright\arraybackslash}p{3cm}
|cccc
|cccc
|cccc
|cccc|
}
\hline

&
& \multicolumn{4}{c|}{\textbf{Vehicle 1}}
& \multicolumn{4}{c|}{\textbf{Vehicle 2}}
& \multicolumn{4}{c|}{\textbf{Vehicle 3}}
& \multicolumn{4}{c|}{\textbf{Vehicle 4}} \\
\cline{3-18}

&
& ADE$\downarrow$ & FDE$\downarrow$ & MR$\downarrow$ & TSR$_{0.5}\uparrow$
& ADE$\downarrow$ & FDE$\downarrow$ & MR$\downarrow$ & TSR$_{0.5}\uparrow$
& ADE$\downarrow$ & FDE$\downarrow$ & MR$\downarrow$ & TSR$_{0.5}\uparrow$
& ADE$\downarrow$ & FDE$\downarrow$ & MR$\downarrow$ & TSR$_{0.5}\uparrow$ \\
\hline

\multirow{7}{*}{\rotatebox{90}{\textbf{w/o Occ.}}}
& Van        & 1.033 & 2.307 & - & 29.06 & 1.040 & 2.415 & - & 26.65 & 1.084 & 2.456 & - & 28.44 & 1.052 & 2.362 & - & 29.59 \\
& Car        & 1.183 & 2.680 & - & 32.66 & 1.197 & 2.721 & - & 33.44 & 1.190 & 2.760 & - & 31.97 & 1.183 & 2.722 & - & 32.92 \\
& Motorcycle & 0.998 & 2.045 & - & 35.02 & 0.979 & 2.111 & - & 36.48 & 1.057 & 2.263 & - & 33.33 & 1.111 & 2.385 & - & 34.80 \\
& Truck      & 0.856 & 2.018 & - & 34.48 & 0.845 & 2.041 & - & 36.61 & 0.833 & 1.948 & - & 33.79 & 0.787 & 1.884 & - & 35.18 \\
& Cyclist    & 0.606 & 1.370 & - & 35.20 & 0.539 & 1.243 & - & 34.24 & 0.578 & 1.312 & - & 38.45 & 0.612 & 1.402 & - & 35.32 \\
& Pedestrian & 0.317 & 0.672 & - & 52.78 & 0.334 & 0.706 & - & 51.55 & 0.338 & 0.719 & - & 52.05 & 0.364 & 0.780 & - & 49.98 \\ \cline{2-18}
& \textbf{Overall} & 0.982 & 2.203 & - & 37.65 & 0.993 & 2.267 & - & 37.55 & 1.006 & 2.298 & - & 36.93 & 1.025 & 2.334 & - & 36.82 \\
\hline

\multirow{12}{*}{\rotatebox{90}{\textbf{under Occ.}}}
& Van        
  & \makecell{1.050 \\ \inc{+0.017}} & \makecell{2.316 \\ \inc{+0.009}} & 11.34 & \makecell{24.49 \\ \decl{-4.57\%}}  
  & \makecell{1.026 \\ \dec{-0.014}} & \makecell{2.362 \\ \dec{-0.053}} & 7.78  & \makecell{22.88 \\ \decl{-3.77\%}} 
  & \makecell{1.077 \\ \dec{-0.007}} & \makecell{2.412 \\ \dec{-0.044}} & 10.24 & \makecell{24.93 \\ \decl{-3.51\%}} 
  & \makecell{1.049 \\ \dec{-0.003}} & \makecell{2.339 \\ \dec{-0.023}} & 8.73  & \makecell{26.58 \\ \decl{-3.01\%}} \\

& Car        
  & \makecell{1.250 \\ \inc{+0.067}} & \makecell{2.788 \\ \inc{+0.108}} & 13.23 & \makecell{26.73 \\ \decl{-5.93\%}} 
  & \makecell{1.243 \\ \inc{+0.046}} & \makecell{2.820 \\ \inc{+0.099}} & 11.21 & \makecell{28.30 \\ \decl{-5.14\%}} 
  & \makecell{1.247 \\ \inc{+0.057}} & \makecell{2.845 \\ \inc{+0.085}} & 9.73  & \makecell{27.47 \\ \decl{-4.50\%}} 
  & \makecell{1.234 \\ \inc{+0.051}} & \makecell{2.808 \\ \inc{+0.086}} & 9.55  & \makecell{28.40 \\ \decl{-4.52\%}} \\

& Motorcycle 
  & \makecell{0.845 \\ \dec{-0.153}} & \makecell{1.762 \\ \dec{-0.283}} & 17.65 & \makecell{27.85 \\ \decl{-7.17\%}}  
  & \makecell{0.875 \\ \dec{-0.104}} & \makecell{1.832 \\ \dec{-0.279}} & 21.37 & \makecell{27.02 \\ \decl{-9.46\%}} 
  & \makecell{0.959 \\ \dec{-0.098}} & \makecell{2.077 \\ \dec{-0.186}} & 22.04 & \makecell{23.60 \\ \decl{-9.73\%}} 
  & \makecell{1.005 \\ \dec{-0.106}} & \makecell{2.153 \\ \dec{-0.232}} & 20.05 & \makecell{25.90 \\ \decl{-8.90\%}} \\

& Truck      
  & \makecell{0.811 \\ \dec{-0.045}} & \makecell{1.867 \\ \dec{-0.178}} & 11.92 & \makecell{28.96 \\ \decl{-5.52\%}} 
  & \makecell{0.772 \\ \dec{-0.073}} & \makecell{1.791 \\ \dec{-0.250}} & 10.82 & \makecell{33.66 \\ \decl{-2.95\%}} 
  & \makecell{0.878 \\ \inc{+0.045}} & \makecell{2.027 \\ \inc{+0.079}} & 3.68  & \makecell{30.11 \\ \decl{-3.68\%}} 
  & \makecell{0.804 \\ \inc{+0.017}} & \makecell{1.902 \\ \inc{+0.018}} & 4.86  & \makecell{32.98 \\ \decl{-2.20\%}} \\

& Cyclist    
  & \makecell{0.524 \\ \dec{-0.082}} & \makecell{1.194 \\ \dec{-0.176}} & 14.54 & \makecell{30.51 \\ \decl{-4.76\%}} 
  & \makecell{0.515 \\ \dec{-0.024}} & \makecell{1.173 \\ \dec{-0.070}} & 16.49 & \makecell{27.67 \\ \decl{-6.57\%}} 
  & \makecell{0.552 \\ \dec{-0.026}} & \makecell{1.235 \\ \dec{-0.077}} & 15.50 & \makecell{30.03 \\ \decl{-8.42\%}} 
  & \makecell{0.577 \\ \dec{-0.035}} & \makecell{1.301 \\ \dec{-0.101}} & 14.55 & \makecell{27.68 \\ \decl{-7.64\%}} \\

& Pedestrian 
  & \makecell{0.354 \\ \inc{+0.037}} & \makecell{0.750 \\ \inc{+0.078}} & 21.04 & \makecell{42.37 \\ \decl{-10.41\%}} 
  & \makecell{0.362 \\ \inc{+0.028}} & \makecell{0.762 \\ \inc{+0.056}} & 22.05 & \makecell{41.55 \\ \decl{-10.00\%}} 
  & \makecell{0.369 \\ \inc{+0.031}} & \makecell{0.778 \\ \inc{+0.059}} & 24.14 & \makecell{40.04 \\ \decl{-12.01\%}} 
  & \makecell{0.395 \\ \inc{+0.031}} & \makecell{0.840 \\ \inc{+0.060}} & 26.80 & \makecell{38.16 \\ \decl{-11.82\%}} \\ \cline{2-18}

& \textbf{Overall} 
  & \makecell{1.043 \\ \inc{+0.061}} & \makecell{2.310 \\ \inc{+0.107}} & 15.17 & \makecell{30.73 \\ \decl{-6.92\%}} 
  & \makecell{1.048 \\ \inc{+0.055}} & \makecell{2.378 \\ \inc{+0.111}} & 14.17 & \makecell{31.28 \\ \decl{-6.27\%}} 
  & \makecell{1.062 \\ \inc{+0.056}} & \makecell{2.397 \\ \inc{+0.099}} & 13.48 & \makecell{30.38 \\ \decl{-6.55\%}} 
  & \makecell{1.083 \\ \inc{+0.058}} & \makecell{2.446 \\ \inc{+0.112}} & 13.63 & \makecell{30.64 \\ \decl{-6.18\%}} \\
\hline

\multirow{16}{*}{\rotatebox{90}{\textbf{Fusion}}}
& Van        
  & \makecell{0.993 \\ \dec{-0.057}} & \makecell{2.189 \\ \dec{-0.127}} & \makecell{10.37 \\ \dec{-0.97\%}} & \makecell{24.82 \\ \incl{+0.33\%}} 
  & \makecell{0.998 \\ \dec{-0.028}} & \makecell{2.300 \\ \dec{-0.062}} & \makecell{7.04 \\ \dec{-0.74\%}} & \makecell{23.15 \\ \incl{+0.27\%}}
  & \makecell{0.981 \\ \dec{-0.096}} & \makecell{2.199 \\ \dec{-0.213}} & \makecell{9.66 \\ \dec{-0.58\%}} & \makecell{25.23 \\ \incl{+0.30\%}}
  & \makecell{1.005 \\ \dec{-0.044}} & \makecell{2.244 \\ \dec{-0.095}} & \makecell{8.05 \\ \dec{-0.68\%}} & \makecell{26.94 \\ \incl{+0.36\%}} \\

& Car        
  & \makecell{1.243 \\ \dec{-0.007}} & \makecell{2.778 \\ \dec{-0.010}} & \makecell{11.92 \\ \dec{-1.31\%}} & \makecell{27.20 \\ \incl{+0.47\%}} 
  & \makecell{1.236 \\ \dec{-0.007}} & \makecell{2.807 \\ \dec{-0.013}} & \makecell{10.42 \\ \dec{-0.79\%}} & \makecell{28.54 \\ \incl{+0.24\%}}
  & \makecell{1.243 \\ \dec{-0.004}} & \makecell{2.842 \\ \dec{-0.003}} & \makecell{8.91  \\ \dec{-0.82\%}} & \makecell{27.87 \\ \incl{+0.40\%}}
  & \makecell{1.232 \\ \dec{-0.002}} & \makecell{2.808 \\ \nog}        & \makecell{8.64  \\ \dec{-0.91\%}} & \makecell{28.66 \\ \incl{+0.26\%}} \\

& Motorcycle 
  & \makecell{0.865 \\ \inc{+0.020}}& \makecell{1.812 \\ \inc{+0.050}} & \makecell{16.53 \\ \dec{-1.12\%}} & \makecell{28.19 \\ \incl{+0.34\%}} 
  & \makecell{0.885 \\ \inc{+0.010}}& \makecell{1.851 \\ \inc{+0.019}} & \makecell{19.61 \\ \dec{-1.76\%}} & \makecell{28.25 \\ \incl{+1.23\%}}
  & \makecell{0.964 \\ \inc{+0.005}}& \makecell{2.087 \\ \inc{+0.010}} & \makecell{20.39 \\ \dec{-1.65\%}} & \makecell{24.50 \\ \incl{+0.90\%}}
  & \makecell{1.036 \\ \inc{+0.031}}& \makecell{2.235 \\ \inc{+0.082}} & \makecell{17.08 \\ \dec{-2.97\%}} & \makecell{26.52 \\ \incl{+0.62\%}} \\

& Truck      
  & \makecell{0.791 \\ \dec{-0.020}}& \makecell{1.824 \\ \dec{-0.043}} & \makecell{11.66 \\ \dec{-0.26\%}} & \makecell{29.27 \\ \incl{+0.31\%}} 
  & \makecell{0.757 \\ \dec{-0.015}}& \makecell{1.758 \\ \dec{-0.033}} & \makecell{10.21 \\ \dec{-0.61\%}} & \makecell{33.79 \\ \incl{+0.13\%}}
  & \makecell{0.839 \\ \dec{-0.039}}& \makecell{1.943 \\ \dec{-0.084}} & \makecell{3.67  \\ \dec{-0.01\%}} & \makecell{30.41 \\ \incl{+0.30\%}}
  & \makecell{0.751 \\ \dec{-0.053}}& \makecell{1.777 \\ \dec{-0.125}} & \makecell{4.69  \\ \dec{-0.17\%}} & \makecell{33.17 \\ \incl{+0.19\%}} \\

& Cyclist    
  & \makecell{0.526 \\ \inc{+0.002}}& \makecell{1.200 \\ \inc{+0.006}} & \makecell{14.23 \\ \dec{-0.31\%}} & \makecell{30.66 \\ \incl{+0.15\%}} 
  & \makecell{0.517 \\ \inc{+0.002}}& \makecell{1.178 \\ \inc{+0.005}} & \makecell{15.96 \\ \dec{-0.53\%}} & \makecell{27.83 \\ \incl{+0.16\%}}
  & \makecell{0.554 \\ \inc{+0.002}}& \makecell{1.239 \\ \inc{+0.004}} & \makecell{14.63 \\ \dec{-0.87\%}} & \makecell{30.47 \\ \incl{+0.44\%}}
  & \makecell{0.584 \\ \inc{+0.007}}& \makecell{1.322 \\ \inc{+0.021}} & \makecell{12.94 \\ \dec{-1.61\%}} & \makecell{28.20 \\ \incl{+0.52\%}} \\

& Pedestrian 
  & \makecell{0.357 \\ \inc{+0.003}} & \makecell{0.759 \\ \inc{+0.009}} & \makecell{19.37 \\ \dec{-1.67\%}} & \makecell{43.04 \\ \incl{+0.67\%}} 
  & \makecell{0.370 \\ \inc{+0.008}} & \makecell{0.781 \\ \inc{+0.019}} & \makecell{20.39 \\ \dec{-1.66\%}} & \makecell{42.10 \\ \incl{+0.55\%}}
  & \makecell{0.375 \\ \inc{+0.006}} & \makecell{0.792 \\ \inc{+0.014}} & \makecell{22.12 \\ \dec{-2.02\%}} & \makecell{40.56 \\ \incl{+0.52\%}}
  & \makecell{0.404 \\ \inc{+0.009}} & \makecell{0.860 \\ \inc{+0.020}} & \makecell{24.67 \\ \dec{-2.13\%}} & \makecell{38.60 \\ \incl{+0.44\%}} \\ \cline{2-18}

& \textbf{Cat L ego}
             & 1.313 & 2.759 & & 
             & 1.274 & 2.764 & & 
             & 1.292 & 2.703 & &  
             & 1.338 & 2.852 & &  \\

& \textbf{Cat L fused}
             & \makecell{1.282  \\ \dec{-0.031}} & \makecell{2.708  \\ \dec{-0.051}} & & 
             & \makecell{1.266 \\ \dec{-0.008}} & \makecell{2.740 \\ \dec{-0.024}} & &
             & \makecell{1.261 \\ \dec{-0.031}} & \makecell{2.648 \\ \dec{-0.055}} & & 
             & \makecell{1.315 \\ \dec{-0.023}} & \makecell{2.809 \\ \dec{-0.043}} & &  \\

& \textbf{Cat S}
             & 0.961 & 2.239 & & 
             & 1.038 & 2.410 & & 
             & 0.952 & 2.258 & &  
             & 1.128 & 2.737 & & \\

& \textbf{Overall (cat L\&S)}
             & \makecell{1.039 \\ \dec{-0.004}} & \makecell{2.305 \\ \dec{-0.005}} & \makecell{13.92 \\ \dec{-1.25\%}} & \makecell{31.21 \\ \incl{+0.48\%}} 
             & \makecell{1.048 \\ \dec{-0.004}} & \makecell{2.380 \\ \dec{-0.012}} & \makecell{13.16 \\ \dec{-1.01\%}} & \makecell{31.63 \\ \incl{+0.35\%}}
             & \makecell{1.058 \\ \dec{-0.004}} & \makecell{2.393 \\ \dec{-0.004}} & \makecell{12.42 \\ \dec{-1.06\%}} & \makecell{30.82 \\ \incl{+0.44\%}}
             & \makecell{1.081 \\ \dec{-0.002}} & \makecell{2.447 \\ \inc{+0.001}} & \makecell{12.39 \\ \dec{-1.24\%}} & \makecell{30.96 \\ \incl{+0.32\%}} \\
\hline
\end{tabular}}
\caption{
DeepAccident per-class results for four vehicles on the validation subsets. All vehicles run the same LSTM-based predictor (keep\_track = 10, past\_trajectory = 10, future\_trajectory = 20 under fps 10Hz). For \textit{under Occ.}, the second line shows the change relative to \textit{w/o Occ.}. For \textit{Fusion} rows, the second line shows the change relative to \textit{under Occ.} rows. \textit{Cat L ego} and \textit{Cat L fused} summarize errors for agents observed by the ego before and after fusion, \textit{Cat S} covers agents only visible to collaborators, and \textit{Overall (cat L\&S)} aggregates both categories. ADE/FDE deltas are absolute, MR and TSR$_{0.5}$ are relative percentage changes (\%).
}
\label{tab:deepaccident_table}
\end{table*}

\begin{table*}[!t]
\centering
\setlength{\tabcolsep}{5pt}
\renewcommand{\arraystretch}{1.2}

\resizebox{\textwidth}{!}{%
\begin{tabular}{
|>{\raggedright\arraybackslash}p{2.4cm}
|cccc
|cccc
|cccc
|cccc|
}
\hline
\multicolumn{1}{|c|}{}
& \multicolumn{8}{c|}{\textbf{Valid subset}}
& \multicolumn{8}{c|}{\textbf{Test subset}} \\
\hline

\multicolumn{1}{|c|}{}
& \multicolumn{4}{c|}{\textbf{Vehicle 1}}
& \multicolumn{4}{c|}{\textbf{Vehicle 2}}
& \multicolumn{4}{c|}{\textbf{Vehicle 1}}
& \multicolumn{4}{c|}{\textbf{Vehicle 2}} \\
\cline{2-17}

\multicolumn{1}{|c|}{}
& ADE$\downarrow$ & FDE$\downarrow$ & MR$\downarrow$ & TSR$_{0.5}\uparrow$ 
& ADE$\downarrow$ & FDE$\downarrow$ & MR$\downarrow$ & TSR$_{0.5}\uparrow$
& ADE$\downarrow$ & FDE$\downarrow$ & MR$\downarrow$ & TSR$_{0.5}\uparrow$ 
& ADE$\downarrow$ & FDE$\downarrow$ & MR$\downarrow$ & TSR$_{0.5}\uparrow$ \\
\hline

\makecell[l]{\textbf{No fusion} \\ \textbf{(w/o Occ.)}}          
  & 0.694 & 1.681 & -     & 55.17 
  & 0.812 & 1.926 & -     & 56.92 
  & 1.245 & 3.023 & -     & 34.36   
  & 1.175 & 2.862 & -     & 35.69  \\

\textbf{No fusion (under Occ.)}        
  & \makecell{0.773 \\ \inc{+0.079}} & \makecell{1.849 \\ \inc{+0.168}}  & 30.42 & \makecell{36.49 \\ \decl{-18.68\%}}  
  & \makecell{0.894 \\ \inc{+0.082}} & \makecell{2.062 \\ \inc{+0.136}} & 34.69 & \makecell{33.68 \\ \decl{-23.24\%}} 
  & \makecell{1.350 \\ \inc{+0.105}} & \makecell{3.202 \\ \inc{+0.179}} & 21.97 & \makecell{24.72 \\ \decl{-9.64\%}} 
  & \makecell{1.259 \\ \inc{+0.084}} & \makecell{2.982 \\ \inc{+0.100}} & 24.21 & \makecell{25.91 \\ \decl{-9.78\%}}   \\
\hline

\textbf{1 sh. cat L ego}     
  & 1.330 & 2.811 & & 
  & 1.275 & 2.765 & & 
  & 1.528 & 3.183 & & 
  & 1.820 & 3.999 & &  \\ 

\textbf{1 sh. cat L fused}   
  & \makecell{1.322 \\ \dec{-0.008}} & \makecell{2.802 \\ \dec{-0.009}} & &
  & \makecell{1.271 \\ \dec{-0.004}} & \makecell{2.760 \\ \dec{-0.005}} & &
  & \makecell{1.519 \\ \dec{-0.009}} & \makecell{3.173 \\ \dec{-0.010}} & &
  & \makecell{1.806 \\ \dec{-0.014}} & \makecell{3.982 \\ \dec{-0.017}} & & \\

\textbf{1 sh. cat S}  
  & 0.934 & 2.257 & &
  & 1.026 & 2.594 &  &  
  & 1.380 & 3.014 &  &  
  & 1.171 & 2.811 &  &  \\

\textbf{1 sh. cat L\&S} 
  & \makecell{0.773 \\ \nog} & \makecell{1.849 \\ \nog} 
  & \makecell{29.19 \\ \dec{-1.23\%}} & \makecell{36.98 \\ \incl{+0.49\%}} 
  & \makecell{0.889 \\ \dec{-0.005}} & \makecell{2.058 \\ \dec{-0.004}} 
  & \makecell{33.76 \\ \dec{-0.93\%}} & \makecell{33.83 \\ \incl{+0.15\%}} 
  & \makecell{1.345 \\ \dec{-0.005}} & \makecell{3.189 \\ \dec{-0.013}} 
  & \makecell{19.43 \\ \dec{-2.54\%}} & \makecell{25.06 \\ \incl{+0.34\%}}  
  & \makecell{1.265 \\ \inc{+0.006}} & \makecell{2.996 \\ \inc{+0.014}} 
  & \makecell{21.70 \\ \dec{-2.51\%}} & \makecell{26.86 \\ \incl{+0.95\%}} \\
\hline

\textbf{2 sh. cat L ego}     
  & 1.335 & 2.803 & & 
  & 1.276 & 2.766 & &  
  & 1.513 & 3.148 & & 
  & 1.810 & 3.981 & &   \\

\textbf{2 sh. cat L fused}   
  & \makecell{1.324 \\ \dec{-0.011}} & \makecell{2.790 \\ \dec{-0.013}} & &
  & \makecell{1.266 \\ \dec{-0.010}} & \makecell{2.757 \\ \dec{-0.009}} & &
  & \makecell{1.501 \\ \dec{-0.012}} & \makecell{3.130 \\ \dec{-0.018}} & &
  & \makecell{1.789 \\ \dec{-0.021}} & \makecell{3.957 \\ \dec{-0.024}} & & \\

\textbf{2 sh. cat S}  
  & 0.511 & 1.230 & & 
  & 0.464 & 1.162 & & 
  & 1.392 & 3.198 & &
  & 1.268 & 2.853 & &  \\

\textbf{2 sh. cat L\&S} 
  & \makecell{0.766 \\ \dec{-0.007}} & \makecell{1.834 \\ \dec{-0.015}} 
  & \makecell{27.76 \\ \dec{-2.66\%}} & \makecell{38.20 \\ \incl{+1.71\%}} 
  & \makecell{0.883 \\ \dec{-0.011}} & \makecell{2.045 \\ \dec{-0.017}} 
  & \makecell{31.89 \\ \dec{-2.80\%}} & \makecell{35.38 \\ \incl{+1.70\%}}
  & \makecell{1.343 \\ \dec{-0.007}} & \makecell{3.189 \\ \dec{-0.013}} 
  & \makecell{18.97 \\ \dec{-3.00\%}} & \makecell{25.11 \\ \incl{+0.39\%}}
  & \makecell{1.268 \\ \inc{+0.009}} & \makecell{2.999 \\ \inc{+0.017}} 
  & \makecell{21.01 \\ \dec{-3.20\%}} & \makecell{26.95 \\ \incl{+1.04\%}} \\
\hline

\textbf{3 sh. cat L ego}     
  & 1.335 & 2.803 & & 
  & 1.276 & 2.766 & & 
  & 1.508 & 3.138 & &
  & 1.818 & 3.986 &  & \\

\textbf{3 sh. cat L fused}   
  & \makecell{1.324 \\ \dec{-0.011}} & \makecell{2.790 \\ \dec{-0.013}} & &
  & \makecell{1.267 \\ \dec{-0.009}} & \makecell{2.759  \\ \dec{-0.007}} & &
  & \makecell{1.496 \\ \dec{-0.012}} & \makecell{3.120 \\ \dec{-0.018}} & & 
  & \makecell{1.791 \\ \dec{-0.020}} & \makecell{3.962 \\ \dec{-0.024}} & & \\

\textbf{3 sh. cat S}  
  & 0.512 & 1.248 & &
  & 0.432 & 1.029 & &
  & 1.403 & 3.216 & & 
  & 1.248 & 2.819 & & \\

\textbf{3 sh. cat L\&S} 
  & \makecell{0.766 \\ \dec{-0.007}} & \makecell{1.835 \\ \dec{-0.014}} 
  & \makecell{27.74 \\ \dec{-2.68\%}} & \makecell{38.21 \\ \incl{+1.72\%}} 
  & \makecell{0.882 \\ \dec{-0.012}} & \makecell{2.042 \\ \dec{-0.020}} 
  & \makecell{31.56 \\ \dec{-3.13\%}} & \makecell{35.68 \\ \incl{+2.00\%}}
  & \makecell{1.343 \\ \dec{-0.007}} & \makecell{3.189 \\ \dec{-0.013}} 
  & \makecell{18.93 \\ \dec{-3.04\%}} & \makecell{25.11 \\ \incl{+0.39\%}}
  & \makecell{1.268 \\ \inc{+0.009}} & \makecell{2.998 \\ \inc{+0.016}} 
  & \makecell{20.98 \\ \dec{-3.23\%}} & \makecell{26.96 \\ \incl{+1.05\%}} \\
\hline

\textbf{4 sh. cat L ego}     
  & 1.344 & 2.824 & &  
  & 1.280 & 2.776 & & 
  & 1.500 & 3.116 & & 
  & 1.811 & 3.986 & & \\

\textbf{4 sh. cat L fused}   
  & \makecell{1.332 \\ \dec{-0.012}}  & \makecell{2.809 \\ \dec{-0.015}}  & &
  & \makecell{1.267 \\ \dec{-0.013}} & \makecell{2.764 \\ \dec{-0.012}} & &
  & \makecell{1.488 \\ \dec{-0.012}} & \makecell{3.098 \\ \dec{-0.018}} & & 
  & \makecell{1.791 \\ \dec{-0.020}} & \makecell{3.962 \\ \dec{-0.024}} & & \\

\textbf{4 sh. cat S}  
  & 0.443 & 1.050 & & 
  & 0.465 & 1.114 & &
  & 1.578 & 3.757 & & 
  & 1.248 & 2.821 & & \\

\textbf{4 sh. cat L\&S} 
  & \makecell{0.763 \\ \dec{-0.010}} & \makecell{1.827  \\ \dec{-0.022}} 
  & \makecell{27.08 \\ \dec{-3.34\%}}  & \makecell{38.82 \\ \incl{+2.33\%}}
  & \makecell{0.876 \\ \dec{-0.018}} & \makecell{2.031 \\ \dec{-0.031}}
  & \makecell{29.68 \\ \dec{-5.01\%}} & \makecell{36.80 \\ \incl{+3.12\%}}
  & \makecell{1.348 \\ \dec{-0.002}} & \makecell{3.204 \\ \inc{+0.002}}
  & \makecell{18.90 \\ \dec{-3.07\%}} & \makecell{25.11 \\ \incl{+0.39\%}}
  & \makecell{1.268 \\ \inc{+0.009}} & \makecell{2.998 \\ \inc{+0.016}}
  & \makecell{20.98 \\ \dec{-3.23\%}} & \makecell{26.96 \\ \incl{+1.05\%}} \\
\hline

\textbf{5 sh. cat L ego}     
  & 1.344 & 2.824 & &
  & 1.280 & 2.776 & &
  & 1.500 & 3.116 & &
  & 1.818 & 3.986 & & \\

\textbf{5 sh. cat L fused}   
  & \makecell{1.331 \\ \dec{-0.013}} &  \makecell{2.808  \\ \dec{-0.016}}  & &
  & \makecell{1.267 \\ \dec{-0.013}} & \makecell{2.764 \\ \dec{-0.012}} & &
  & \makecell{1.488 \\ \dec{-0.012}} & \makecell{3.098 \\ \dec{-0.018}} & &
  & \makecell{1.791 \\ \dec{-0.027}} & \makecell{3.962 \\ \dec{-0.024}} & & \\

\textbf{5 sh. cat S}  
  & 0.444 & 1.051 & &
  & 0.465 & 1.114 & &
  & 1.578 & 3.757 & &
  & 1.245 & 2.816 & & \\

\textbf{5 sh. cat L\&S} 
  & \makecell{0.763 \\ \dec{-0.010}} & \makecell{1.827  \\ \dec{-0.022}} 
  & \makecell{27.08 \\ \dec{-3.34\%}}  & \makecell{38.82 \\ \incl{+2.33\%}}
  & \makecell{0.876 \\ \dec{-0.018}} & \makecell{2.031 \\ \dec{-0.031}}
  & \makecell{29.68 \\ \dec{-5.01\%}} & \makecell{36.80 \\ \incl{+3.12\%}}
  & \makecell{1.348 \\ \dec{-0.002}} & \makecell{3.204 \\ \inc{+0.002}}
  & \makecell{18.90 \\ \dec{-3.07\%}} & \makecell{25.11 \\ \incl{+0.39\%}}
  & \makecell{1.268 \\ \inc{+0.009}} & \makecell{2.998 \\ \inc{+0.016}}
  & \makecell{20.98 \\ \dec{-3.23\%}} & \makecell{26.96 \\ \incl{+1.05\%}} \\
\hline

\textbf{6 sh. cat L ego}     
  & 1.344 & 2.824 & &
  & 1.280 & 2.776 & &
  & 1.500 & 3.116 & &
  & 1.811 & 3.986 & & \\

\textbf{6 sh. cat L fused}   
  & \makecell{1.327 \\ \dec{-0.017}} & \makecell{2.803 \\ \dec{-0.021}} & &
  & \makecell{1.267 \\ \dec{-0.013}} & \makecell{2.765 \\ \dec{-0.011}} & &
  & \makecell{1.488 \\ \dec{-0.012}} & \makecell{3.098 \\ \dec{-0.018}} & &
  & \makecell{1.791 \\ \dec{-0.012}} & \makecell{3.962 \\ \dec{-0.024}} & & \\

\textbf{6 sh. cat S}  
  & 0.412 & 0.974 & &
  & 0.465 & 1.113 & & 
  & 1.578 & 3.757 & &
  & 1.248 & 2.821 & & \\

\textbf{6 sh. cat L\&S} 
  & \makecell{0.760 \\ \dec{-0.013}} &  \makecell{1.819 \\ \dec{-0.030}} 
  & \makecell{26.77 \\ \dec{-3.65\%}} & \makecell{39.09 \\ \incl{+2.6\%}} 
  & \makecell{0.875 \\ \dec{-0.019}} & \makecell{2.030 \\ \dec{-0.032}}
  & \makecell{29.56 \\ \dec{-5.13\%}} & \makecell{36.90 \\ \incl{+3.22\%}}
  & \makecell{1.348 \\ \dec{-0.002}} & \makecell{3.204 \\ \inc{+0.002}}
  & \makecell{18.90 \\ \dec{-3.07\%}} & \makecell{25.11 \\ \incl{+0.39\%}}
  & \makecell{1.268 \\ \inc{+0.009}} & \makecell{2.998 \\ \inc{+0.016}}
  & \makecell{20.98 \\ \dec{-3.23\%}} & \makecell{26.96 \\ \incl{+1.05\%}} \\
\hline

\end{tabular}}
\caption{
OPV2V dataset ablation results by node category and number of collaborating vehicles. All vehicles run the same Transformer predictor with $1\,\mathrm{s}$ history and $2\,\mathrm{s}$ horizon (\textit{tr12}), where the broadcasting vehicle use a shorter tracklet lifetime than the aggregating vehicle. For each number of sharing vehicles (1--6), \emph{cat L ego} and \emph{cat L fused} report errors for agents observed by the ego, before and after fusion; \emph{cat S} covers agents only visible to collaborators; and \emph{cat L\&S} shows overall performance across both categories. ADE/FDE deltas are absolute, MR and TSR$_{0.5}$ are relative percentage changes (\%).
}

\label{tab:opv2v_table_2}
\end{table*}

To evaluate collaborative trajectory prediction, three groups of experiments are conducted. 
Section~\ref{sec:indiv_collab} compares individual and collaborative prediction on two datasets under different predictor configurations, and varies the number of collaborators to analyze trends across both agent- and node-level categories.  Section~\ref{sec:real_setup} reports results on a real-world dataset using the real-deployment setup and discusses runtime feasibility. Finally, Section~\ref{sec:latency} focuses on communication aspects, examining the communication volume and the impact of message delays on performance.

In the result tables, rows marked \textit{(w/o Occ.)} report single-vehicle performance under perfect detections and tracklets. Rows marked \textit{(under Occ.)} report single-vehicle performance in the presence of occlusions, and the reported deltas are computed relative to the corresponding \textit{(w/o Occ.)} setting. 
All \emph{Fusion} rows are compared against the corresponding \textit{(under Occ.)} baseline, so their deltas are reported relative to that setting. The vehicle, which is evaluated in the aggregating mode is refereed as ego in the experiments. 

Under occlusion, both displacement errors and miss rate are expected to increase compared to the fully observable case. With fusion, the miss rate is expected to decrease; however, ADE/FDE may fluctuate because these errors are computed only over matched targets, and fusion can recover targets that were previously missed and may have larger errors. Thus, TSR is the primary metric for comparing configurations, since it jointly reflects accuracy and coverage. The exact dataset-specific settings and experimental parameters are provided in the \href{https://github.com/AV-Lab/Collab_Late_Trajectory_Prediction}{repository} to ensure full reproducibility.

\subsection{Individual versus Collaborative Modes}
\label{sec:indiv_collab}

\paragraph{\textbf{Effect of predictor configuration}} 
Table~\ref{tab:opv2v_table} reports per-vehicle results on the OPV2V dataset for \textit{Vehicle 1} and \textit{Vehicle 2}, which are present in all scenarios, for both validation and test subsets. The setting \textit{No fusion (w/o Occ.)} is evaluated for four predictor configurations: Transformer and LSTM predictors, each with two temporal settings, \textit{12} (1\,s history, 2\,s prediction horizon) and \textit{23} (2\,s history, 3\,s horizon). Across all configurations, occlusions cause a moderate increase in $\mathrm{ADE}$/$\mathrm{FDE}$ and a strong drop in $\mathrm{TSR}_{0.5}$, with a  maximum drop reaching $23.24\%$ reported in \textit{No fusion (under Occ.)}.

To analyze the cases where aggregating vehicles uses weaker or stronger predictors, we first compared the performance of individual predictors without occlusion. In general, LSTM performs better in displacement errors, especially in the test datasets; for \textit{Vehicle 1} the difference reached roughly 35 and 90~cm in ADE and FDE in comparison with the Transformer in the shorter-horizon setting. Additionally, Transformer displacement errors are consistently more impacted under occlusion than LSTM. However, for the longer horizon, the trend of lower errors for LSTM remains, with a large margin of reduced displacement error but a substantial drop in $\mathrm{TSR}$ in comparison with the Transformer (e.g., 10.10\% versus 56.23\% without occlusion). To further investigate this behavior and eliminate any possibility of error, the success threshold was increased to $\tau = 0.75$. Under this condition, LSTM and Transformer achieve $\mathrm{TSR}_{0.75}$ of $32.78$\% and $34.77$\%, respectively, for \textit{Vehicle 1}, showing close performance. Combined with the lower displacement errors, this indicates that LSTM trajectories are more stable: the error histogram is less skewed and exhibits shorter tails, with most predictions concentrated around the main low-error peak rather than in the tails. In contrast, the Transformer handles certain motion patterns better, which yields good performance when $\tau = 0.5$, but it also produces more large-error outliers, resulting in marginally higher average displacement errors and a more right-skewed error distribution.

In \emph{Fusion} rows each predictors configuration follows the format $ego\_predictor:others\_predictor$, where homogeneous settings use the same model and prediction horizon for both ego and received trajectories (e.g., \textit{tr12:tr12}, \textit{lstm23:lstm23}), and mixed settings combine different predictors or same predictor but different horizon (e.g., \textit{tr12:tr23}, \textit{lstm12:tr12}). Across all configurations, vehicles, and splits, fusion consistently reduces $\mathrm{MR}$ by about $2$--$3.5$\% and increases $\mathrm{TSR}_{0.5}$ in the range of $0.3$--$2$\%, both in homogeneous and mixed settings. Changes in $\mathrm{ADE}$ and $\mathrm{FDE}$ do not follow a strict monotonic trend: when fusion recovers agents that were previously missed, these objects enter the matched set and can slightly increase average displacement errors, even though overall coverage and $\mathrm{TSR}_{0.5}$ improve. A more detailed analysis of this effect is provided in a later table, where the dataset is stratified into category~L and category~S nodes to examine the corresponding displacement error traces.

For the shorter-horizon \textit{12} setting, comparing the homogeneous configuration (\textit{tr12:tr12}) with the mixed configuration where the ego still uses Transformer but the received trajectories come from the stronger LSTM predictor (\textit{tr12:lstm12}), the mixed setting yields a larger reduction in displacement errors in all cases, except for \textit{Vehicle 2} in FDE in valid subset. In the reverse scenario, when the ego uses LSTM and is paired either with LSTM (\textit{lstm12:lstm12}) or Transformer (\textit{lstm12:tr12}), the configuration \textit{lstm12:lstm12} consistently achieves the stronger improvement in displacement errors for both vehicles. 

In regard to different horizons, when the ego uses the longer horizon predictor and broadcasting vehicles use the shorter horizon predictor (\emph{Fusion, \textit{lstm23:lstm12}}) compared to the homogeneous long-horizon configuration (\emph{Fusion, \textit{lstm23:lstm23}}), the performance is tangibly worse in terms of displacement errors; for example, on the test subset for \textit{Vehicle 2}, ADE/FDE increase from $1.522/3.651$ to $1.562/3.846$. The same trend is supported by the comparison between the Transformer configurations \emph{Fusion, \textit{tr23:tr12}} and \emph{Fusion, \textit{tr23:tr23}}. When the ego uses the shorter hroizon LSTM predictor and the other vehicles use the longer horizon (\emph{Fusion, \textit{lstm12:lstm23}}), the FDE shows improvement for all cases, while ADE shows a decrease in comparison with the homogeneous \emph{Fusion, \textit{lstm12:lstm12}} configuration. For the Transformer, the longer-horizon predictor yields consistently worse behavior when the ego uses the shorter horizon predictor, compared to the homogeneous short-horizon configuration (i.e., \emph{Fusion, \textit{tr12:tr23}} versus \emph{Fusion, \textit{tr12:tr12}}). In theory, longer-horizon predictions from other vehicles could provide additional foresight that the ego might exploit, but in practice these configurations did not yield the hypothesised gains, most likely because further tuning and adjustment of the fusion mechanism would be required.

\paragraph{\textbf{Effect of semantic classes and node categories (L/S)}}
Table~\ref{tab:deepaccident_table} shows the results of individual and collaborative experiments on the DeepAccident valid dataset. In this dataset, experiments are run for all four vehicles, which each acting as ego at runtime while other vehicles operate in the broadcasting mode. Results are additionally reported per semantic class (van, car, motorcycle, truck, cyclist, and pedestrian). The overall fusion performance is summarized in the row \emph{Overall (cat L\&S)}. For category~L, the ego-only performance is reported in \emph{Cat L ego}, and \emph{Cat L fused} gives the corresponding errors after fusion for the same set of agents. Consistent with the OPV2V experiments, occlusions increase the missing rate and reduce TSR, whereas late fusion reverses part of this degradation: across the four vehicles, fusion reduces the overall missing rate by about $1.0$--$1.3$\% and increases TSR$_{0.5}$ in the range of $0.3$--$0.5$\%. Regarding displacement errors, ADE decreases after fusion for all vehicles, and FDE also decreases except for \textit{Vehicle 4}.

At the level of semantic classes, for vans, cars, and trucks fusion not only reduces $\mathrm{MR}$ and increases $\mathrm{TSR}_{0.5}$, but also consistently lowers $\mathrm{ADE}$ and $\mathrm{FDE}$ for all vehicles. The largest displacement-error reduction is observed for vans with \textit{Vehicle 3} as ego, with an improvement of almost $0.10$\,m in ADE and $0.21$\,m in FDE. A different trend appears for motorcycles, cyclists, and pedestrians. For motorcycles and cyclists, introducing occlusions reduces displacement errors while increasing the missing rate, effectively pruning some hard-to-predict agents; with fusion, the missing rate decreases again, bringing these agents back and increasing displacement errors. For pedestrians, occlusions increase both ADE and FDE, and fusion further raises them, indicating that pedestrian trajectories are particularly susceptible to occlusion-induced errors.

At the category level, comparing \emph{Cat L ego} to \emph{Cat L fused}, late fusion consistently reduces $\mathrm{ADE}$ by $0.008$–$0.031$\,m and $\mathrm{FDE}$ by $0.024$–$0.055$\,m across vehicles, indicating that the residual GP correction refines already-visible agents but only by a small margin; further gains on these nodes would require additional design improvements. The net change in overall displacement errors then depends on how far the \emph{Cat S} errors lie from the mean ego error for agents on which the ego does not collaborate.

\paragraph{\textbf{Effect of number of collaborators}} 
Finally, Table~\ref{tab:opv2v_table_2} completes the experiments and decomposes OPV2V performance for homogeneous \textit{tr12} predictors by node category and number of sharing vehicles. For the node categories, the other vehicles’ trackers are configured with a shorter tracklet lifetime than the ego (5 versus 10 frames), so under occlusions the shared estimates are intentionally more certain than the ego’s. Comparing \emph{cat L ego} and \emph{cat L fused} shows that fusion consistently reduces $\mathrm{ADE}$ and $\mathrm{FDE}$ for locally observed agents, but only by about $0.01$–$0.02$\,m, i.e., the same small margin as in the previous experiments; this suggests that predictor error may dominate the error introduced by occlusions, limiting the tangible gains from residual correction alone. As for the number of sharing vehicles, for \textit{Vehicle 1} on the validation subset, increasing from one to six sharing vehicles improves TSR$_{0.5}$ by about $2.11$ points, with a similar trend for \textit{Vehicle 2}. On the test subset, however, the best performance is achieved when two vehicles share for \textit{Vehicle 1} and three vehicles share for \textit{Vehicle 2}; adding more collaborators yields no further benefit and only increases fusion overhead and runtime. This raises the question of an optimal number of sharing vehicles, which is left as an interesting direction for future work.

\subsection{V2V4Real Evaluation}
\label{sec:real_setup}

\paragraph{\textbf{Performance}} Tables~\ref{tab:v2v4real_results_veh_1} and \ref{tab:v2v4real_results_veh_2} report the results of individual and collaborative experiments on the V2V4Real dataset for the validation and test subsets of \textit{Vehicle~1} and \textit{Vehicle~2}, under casting-ray (\emph{Casting Ray\,$\rightarrow$\,Tracker (ID asc.)}) and real-deployment setups (\emph{Detector\,$\rightarrow$\,Tracker}). In all V2V4Real dataset experiments, the individual predictor for both vehicles is an LSTM, using 1 second of past trajectory and a 2-second prediction horizon at 10 Hz. In the ray-casting case, occlusions introduce a miss rate of roughly $23$--$29\%$, which is about a factor of two lower than the $45$--$55\%$ miss rates observed with the detector--tracker pipeline, suggesting that ray-casting provides a controlled yet non-trivial proxy for studying performance under occlusions. Across both vehicles and both setups, the trend under collaboration is consistent: late fusion reduces $\mathrm{MR}$ and improves $\mathrm{TSR}_{0.5}$, confirming that multi-vehicle fusion recovers missed agents and increases successful trajectory coverage. Fusion results are averaged over 10 runs and bracketed ranges report $95\%$ confidence intervals.

\begin{table}[!h]
\centering
\scriptsize
\setlength{\tabcolsep}{2.6pt}
\renewcommand{\arraystretch}{1.08}
\resizebox{\columnwidth}{!}{%
\begin{tabular}{|c|c|l|cccc|}
\hline
\multirow{2}{*}{} & \multirow{2}{*}{} & \multirow{2}{*}{\textbf{Setting}}
& \multicolumn{4}{c|}{\textbf{Vehicle 1}} \\
\cline{4-7}
& & &
ADE$\downarrow$ & FDE$\downarrow$ & MR$\downarrow$ & TSR$_{0.5}\uparrow$ \\
\hline

\multirow{10}{*}{\rotatebox{90}{\makecell{Casting Ray$\rightarrow$ \\ Tracker (ID asc.)}}}
& \multirow{5}{*}{\rotatebox{90}{valid}}
& w/o Occ.    & 1.380 & 2.609 & --    & 63.50  \\
& & under Occ.  & \makecell{1.381 \\ \inc{\scriptsize{+0.001}}} & \makecell{2.611 \\ \inc{\scriptsize{+0.002}}} & 23.57 & \makecell{46.67 \\ \decl{\scriptsize{-16.83\%}}}  \\
& & Fusion  & \makecell{1.381 \\ \scriptsize{\nog}} & \makecell{2.611 \\ \scriptsize{\nog}} & \makecell{$23.10,\; [23.09,\, 23.10]$ \\ \dec{\scriptsize{-0.47\%}}}  & \makecell{$47.02,\; [47.02,\, 47.03]$ \\ \incl{\scriptsize{+0.35\%}}} \\
\cline{2-7}
& \multirow{5}{*}{\rotatebox{90}{test}}
& w/o Occ.    & 1.617 & 3.115 & --    & 56.99 \\
& & under Occ.   & \makecell{1.636 \\ \inc{\scriptsize{+0.019}}} & \makecell{3.146 \\ \inc{\scriptsize{+0.031}}} & 28.84 & \makecell{36.27 \\ \decl{\scriptsize{-20.72\%}}}  \\
& & Fusion  & \makecell{1.638 \\  \inc{\scriptsize{+0.002}}} & \makecell{3.144 \\  \inc{\scriptsize{-0.002}}} & \makecell{$25.39\; [25.37,\, 23.40]$ \\ \dec{\scriptsize{-3.45\%}}} & \makecell{$38.69,\; [38.69,\, 38.70]$ \\ \incl{\scriptsize{+2.42\%}}}  \\
\hline

\multirow{6}{*}{\rotatebox{90}{\makecell{Detector$\rightarrow$ \\ Tracker}}}
& \multirow{3}{*}{\rotatebox{90}{valid}}
& Indiv & 1.123 & 2.098 & 45.78 & 37.88  \\
& & Fusion  & \makecell{1.139 \\ \inc{\scriptsize{+0.016}}} & \makecell{2.139 \\ \inc{\scriptsize{+0.041}}} & \makecell{$44.02,\; [44.00,\, 44.03]$ \\ \dec{\scriptsize{-1.76\%}}} & \makecell{$39.09,\; [39.08,\, 39.11]$  \\ \incl{\scriptsize{+1.21\%}}} \\
\cline{2-7}
& \multirow{3}{*}{\rotatebox{90}{test}}
& Indiv & 1.524 & 3.045 & 54.44 & 24.88  \\
& & Fusion  & \makecell{1.517 \\ \dec{\scriptsize{-0.007}}} & \makecell{3.033 \\ \dec{\scriptsize{-0.012}}} & \makecell{$52.77,\; [52.76,\, 52.78]$ \\ \dec{\scriptsize{-1.67\%}}}
 & \makecell{$26.57,\; [26.56,\, 26.58]$ \\ \incl{\scriptsize{+1.69\%}}}
  \\
\hline
\end{tabular}}
\caption{Trajectory prediction on V2V4Real for \textit{Vehicle~1} under ray-casting and detector-based setups, comparing individual and collaborative (fusion) prediction.}
\label{tab:v2v4real_results_veh_1}
\end{table}

\begin{table}[!h]
\centering
\scriptsize
\setlength{\tabcolsep}{2.6pt}
\renewcommand{\arraystretch}{1.08}
\resizebox{\columnwidth}{!}{%
\begin{tabular}{|c|c|l|cccc|}
\hline
\multirow{2}{*}{} & \multirow{2}{*}{} & \multirow{2}{*}{\textbf{Setting}}
& \multicolumn{4}{c|}{\textbf{Vehicle 2}} \\
\cline{4-7}
& & &
ADE$\downarrow$ & FDE$\downarrow$ & MR$\downarrow$ & TSR$_{0.5}\uparrow$ \\
\hline

\multirow{10}{*}{\rotatebox{90}{\makecell{Casting Ray$\rightarrow$ \\ Tracker (ID asc.)}}}
& \multirow{5}{*}{\rotatebox{90}{valid}}
& w/o Occ.    & 1.646 & 3.198 & --    & 56.69  \\
& & under Occ.  & \makecell{1.664 \\ \inc{\scriptsize{+0.020}}} & \makecell{3.228 \\ \inc{\scriptsize{+0.030}}} & 28.74 & \makecell{36.05 \\ \decl{\scriptsize{-20.64\%}}}  \\
& & Fusion  & \makecell{1.667 \\ \inc{\scriptsize{+0.003}}} & \makecell{3.226 \\ \dec{\scriptsize{-0.002}}} & \makecell{$25.31,\; [25.30,\, 25.32]$ \\ \dec{\scriptsize{-3.43\%}}}  & \makecell{$38.46,\; [38.45,\, 38.46]$ \\ \incl{\scriptsize{+2.41\%}}} \\
\cline{2-7}
& \multirow{5}{*}{\rotatebox{90}{test}}
& w/o Occ.    & 1.408 & 2.690 & --    & 63.22 \\
& & under Occ.   & \makecell{1.409 \\ \inc{\scriptsize{+0.001}}} & \makecell{2.692 \\ \inc{\scriptsize{+0.002}}} & 23.49 & \makecell{46.45 \\ \decl{\scriptsize{-16.77\%}}}  \\
& & Fusion  & \makecell{1.409 \\ \scriptsize{\nog}} & \makecell{2.692 \\ \scriptsize{\nog}} & \makecell{$23.03\; [23.02,\, 23.03]$ \\ \dec{\scriptsize{-0.46\%}}} & \makecell{$46.80,\; [46.79,\, 46.81]$ \\ \incl{\scriptsize{+0.35\%}}}  \\
\hline

\multirow{6}{*}{\rotatebox{90}{\makecell{Detector$\rightarrow$ \\ Tracker}}}
& \multirow{3}{*}{\rotatebox{90}{valid}}
& Indiv & 1.149 & 2.342 & 55.29 & 24.77  \\
& & Fusion  & \makecell{1.143 \\ \dec{\scriptsize{-0.006}}} & \makecell{2.330 \\ \dec{\scriptsize{-0.012}}} & \makecell{$53.63,\; [53.62,\, 53.63]$ \\ \dec{\scriptsize{-1.66\%}}} & \makecell{$26.46,\; [26.45,\, 26.46]$  \\ \incl{\scriptsize{+1.69\%}}} \\
\cline{2-7}
& \multirow{3}{*}{\rotatebox{90}{test}}
& Indiv & 1.462 & 2.717 & 45.11 & 37.78  \\
& & Fusion  & \makecell{1.478 \\ \inc{\scriptsize{+0.016}}} & \makecell{2.757 \\ \inc{\scriptsize{+0.040}}} & \makecell{$43.36,\; [43.33,\, 43.38]$ \\ \dec{\scriptsize{-1.75\%}}}
 & \makecell{$39.00,\; [38.98,\, 39.02]$ \\ \incl{\scriptsize{+1.22\%}}}
  \\
\hline
\end{tabular}}
\caption{Trajectory prediction on V2V4Real for \textit{Vehicle~2} under ray-casting and detector-based setups, comparing individual and collaborative (fusion) prediction.}
\label{tab:v2v4real_results_veh_2}
\end{table}

\begin{figure}[h!]
\centering
   \includegraphics[width=0.48\textwidth]{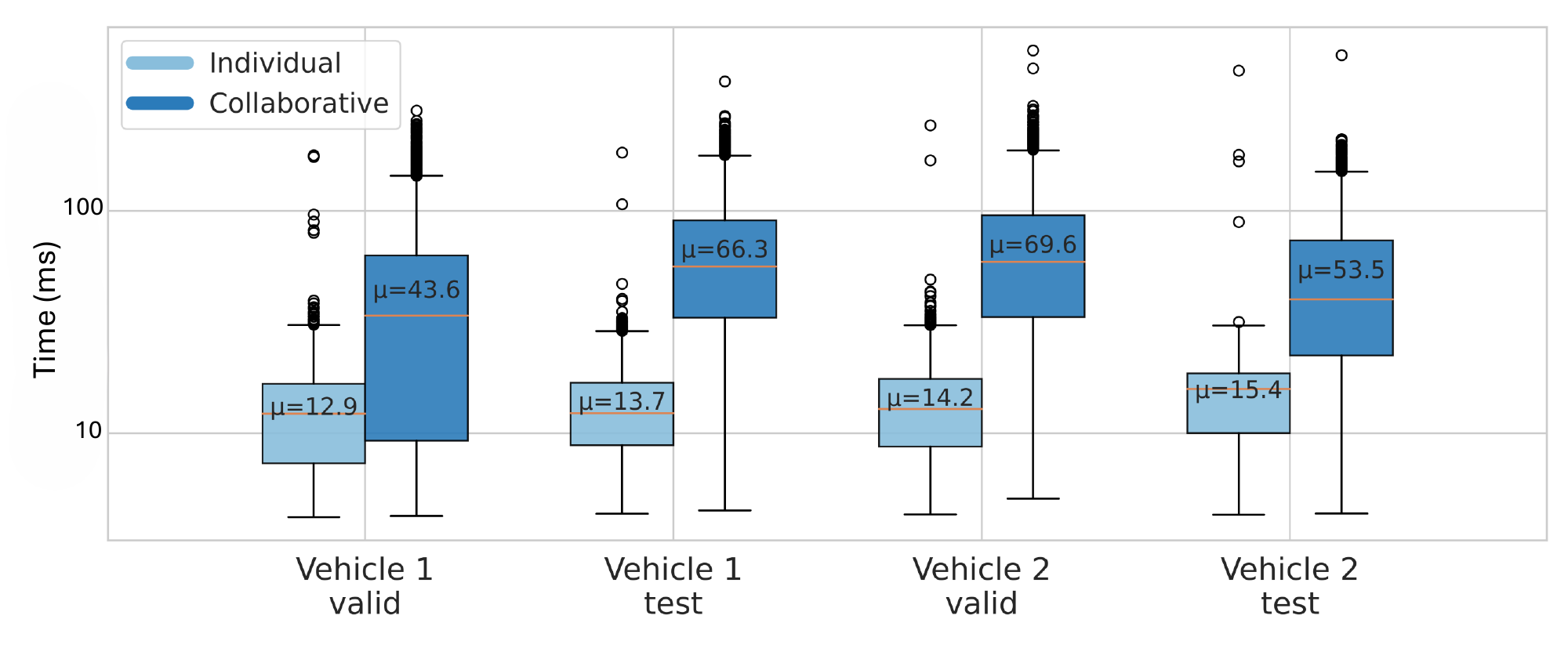}
\caption{Runtime comparison between individual trajectory prediction and collaborative prediction with fusion for V2V4Real dataset.}
   \label{fig:run_time}
\end{figure}

\paragraph{\textbf{Time Feasibility}} Figure~\ref{fig:run_time} compares the runtime of individual and collaborative settings for online deployment. Boxplots show the distribution of per-step runtimes, with ($\mu$) indicating the mean runtime in milliseconds. The individual setting measures only the local predictor, whereas the collaborative setting includes the full fusion cycle (all local predictions plus GP-based refinement). Mean runtimes for individual prediction range from about $12.9$\,ms to $15.4$\,ms, while collaborative prediction increases this by roughly a factor of four ($43.6$–$69.6$\,ms), keeping all runtimes below $100$\,ms. The remaining outliers are caused by frames with many tracked agents and can be suppressed by limiting prediction to agents within a proximity/radius around the ego vehicle. The fusion computation time is governed by the cost of online kernel hyperparameter optimization. Improving runtime can be potentially be done through filtering noisy shared tracks before fusion, and precomputing parts of the GP model so that hyperparameter optimization is run only for those nodes when necessary. This optimization is left for exploration as future work.

\subsection{Communication Volume and Latency}
\label{sec:latency}

\begin{figure}[h!]
\centering
   \includegraphics[width=0.48\textwidth]{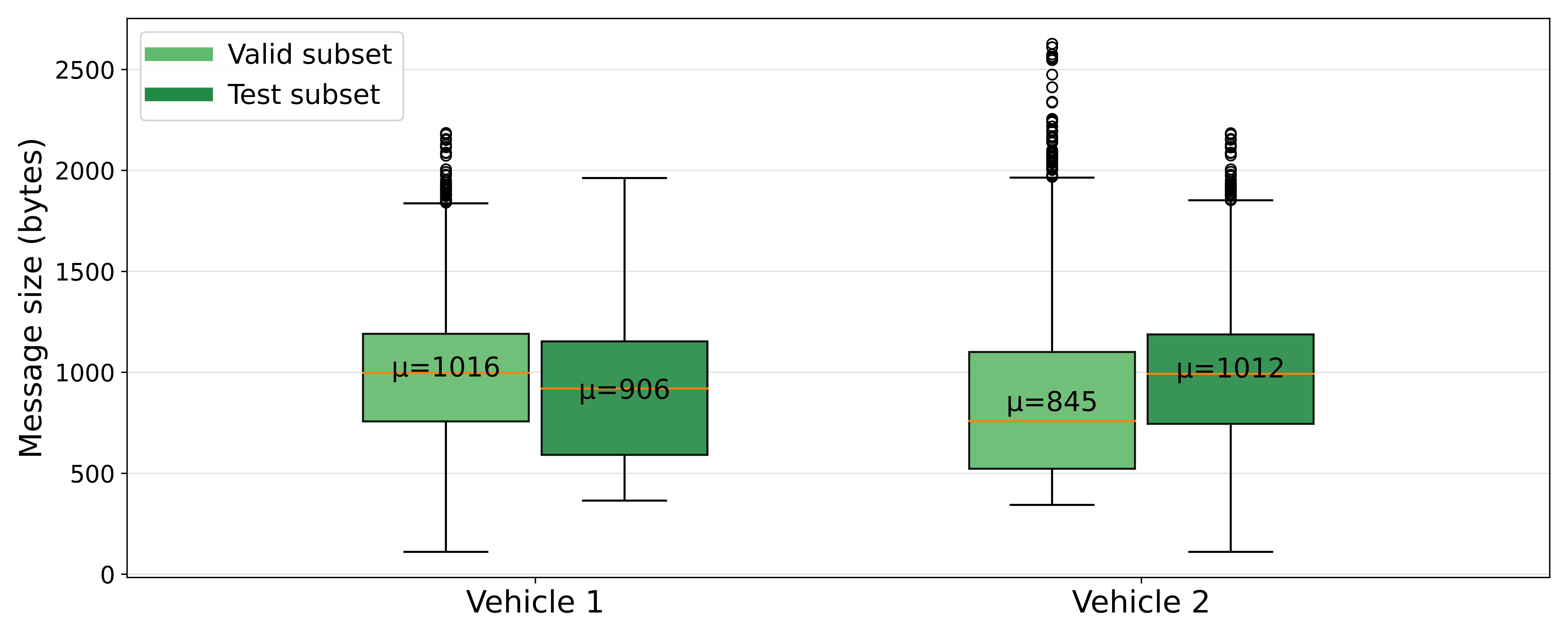}
\caption{Per-broadcast message sizes for Vehicles~1 and~2 on V2V4Real (validation and test subsets).}
   \label{fig:v2v4real_msg_sizes}
\end{figure}

\begin{figure}[h!]
\centering
   \includegraphics[width=0.48\textwidth]{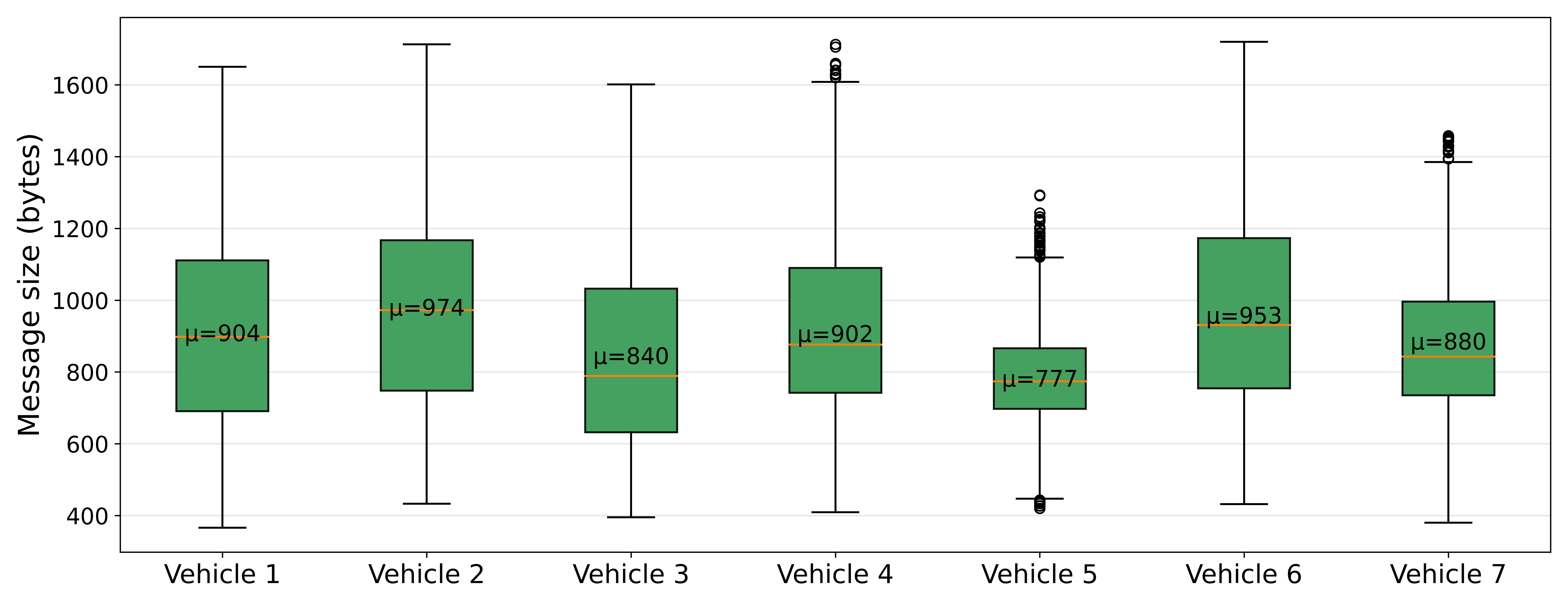}
\caption{Per-broadcast message sizes for all seven vehicles on OPV2V (validation subset).}
   \label{fig:opv2v_valid_msg_sizes}
\end{figure}

\begin{figure}[h!]
\centering
   \includegraphics[width=0.48\textwidth]{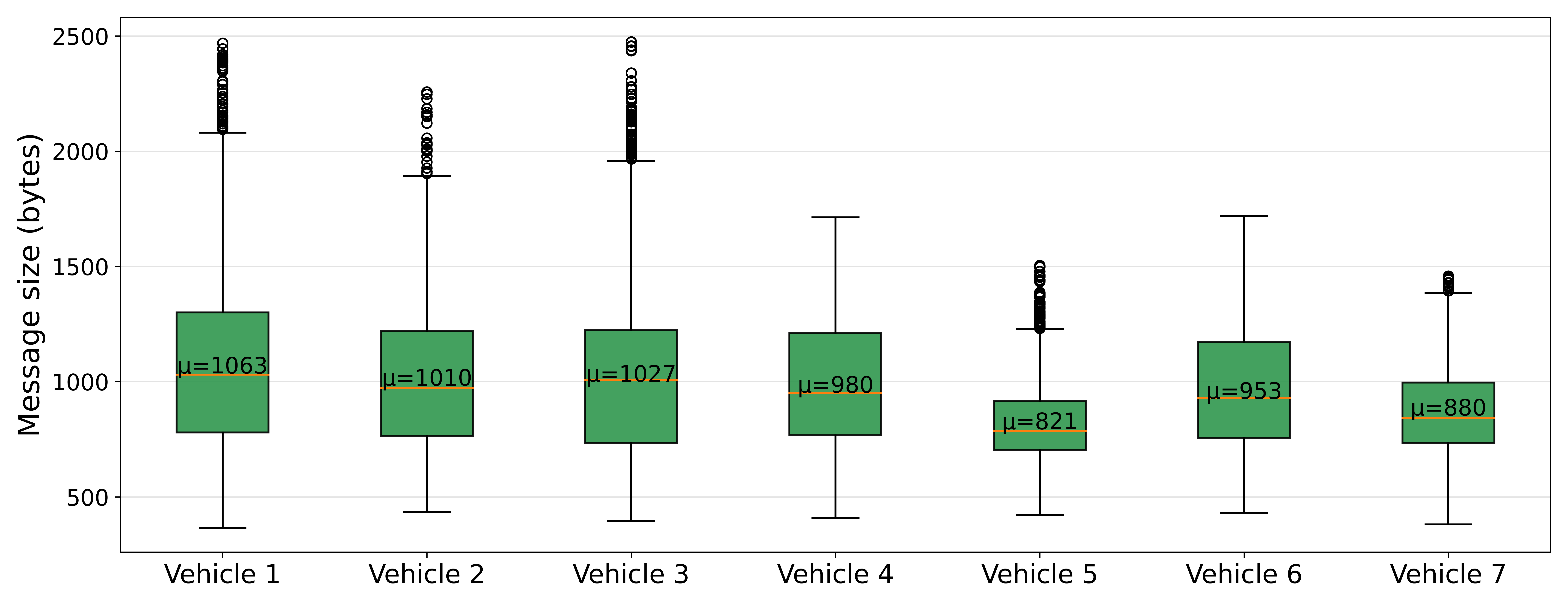}
\caption{Per-broadcast message sizes for all seven vehicles on OPV2V (test subset).}
   \label{fig:opv2v_test_msg_sizes}
\end{figure}

\begin{figure}[h!]
\centering
   \includegraphics[width=0.48\textwidth]{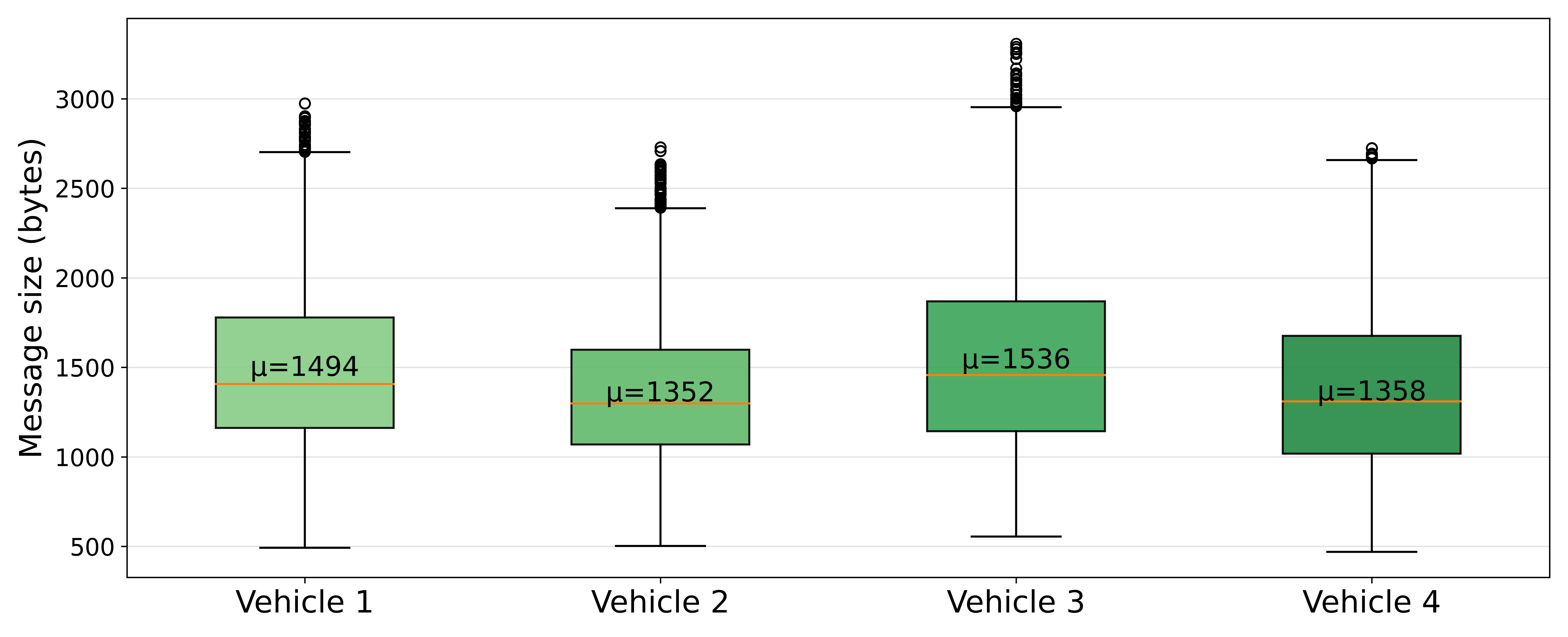}
\caption{Per-broadcast message sizes for four vehicles on DeepAccident (valid subset).}
   \label{fig:deepaccident_msg_sizes}
\end{figure}

\paragraph{\textbf{Communication Volume}} For each dataset, the per-broadcast message-size distribution is plotted for every vehicle, with boxplots indicating the spread and annotated means ($\mu$). Figures~\ref{fig:v2v4real_msg_sizes}, \ref{fig:opv2v_valid_msg_sizes}, and \ref{fig:opv2v_test_msg_sizes} (V2V4Real and OPV2V) show that the mean per-message sizes for all vehicles are below the 1{,}500-byte upper bound imposed by the 5G NR-V2X design, with most of the distribution well within this limit. DeepAccident (Figure~\ref{fig:deepaccident_msg_sizes}) exhibits heavier tails and mean sizes that approach or exceed 1{,}500~bytes, reflecting its higher object density.

One way to lower the message size is to increase the compression of the payload. For compression, we use Zstandard (Zstd), a modern LZ77-based scheme with Finite State Entropy coding and a configurable compression level. Increasing the compression level can further reduce message sizes, but at the cost of higher runtime. As an alternative direction, an optimization scheme could share trajectories only for a subset of targets (e.g., those closest to the ego or most informative), thereby maximizing information gain while keeping communication volume and latency under the threshold. Exploring such adaptive sharing policies remains an interesting direction for future work.

\paragraph{\textbf{Latency}} The simplified latency model is constructed based on the findings reported in \cite{depontemueller2021itsg5cv2x}, which evaluates latency and packet drop for the C-V2X protocol under packet sizes of 200, 800, and 1000 bytes at a 40~Hz broadcast frequency. To adapt these observations to our setting, each transmitted packet of size $b$ bytes is first subjected to a Bernoulli packet-drop model and, if successfully received, is assigned a delay $\Delta$:
\begin{equation*}
D(b)\sim \mathrm{Bernoulli}\bigl(1-p_{\mathrm{drop}}(b)\bigr),
\end{equation*}
\begin{equation*}
\Delta = k\,b + X,\qquad X \sim \mathrm{LogNormal}(\mu,\sigma),
\label{eq:delay_model}
\end{equation*}
where $D(b)=1$ indicates successful packet delivery and $D(b)=0$ indicates packet loss. The term $k\,b$ captures the increase of latency with packet size, while $X$ models right-skewed channel-access and scheduling jitter.

For the log-normal component, $(\mu,\sigma)$ are estimated from the 50th and 95th percentile quantiles of the reported C-V2X latency CDF for 200-byte packets in \cite{depontemueller2021itsg5cv2x}, resulting in $\mu=3.135$ and $\sigma=0.182$. The coefficient $k$ is estimated from the increase in the 50th percentile delay across the three reported packet sizes, yielding $k=0.0197$. Packet drop is modeled as a Bernoulli random event with size-dependent probability, where the values are set from the reported packet error rate CDFs in \cite{depontemueller2021itsg5cv2x}. Accordingly, 200-byte packets are assumed not to be dropped, while larger packets are assigned higher drop probabilities to reflect the reported increase in packet error rate with packet size:
\begin{equation*}
p_{\mathrm{drop}}(b)=
\begin{cases}
0, & b \le 400,\\
0.08, & 400 < b \le 900,\\
0.10, & b > 900.
\end{cases}
\label{eq:drop_model}
\end{equation*}

A packet broadcast at time $t$ becomes available to the receiver at $t+\Delta$. When fusion is performed, the receiver uses only packets that are already available and ignores packets that have not yet arrived until they become available. Results are summarized in Table~\ref{tab:v2v4real_time_delay}.

\begin{table}[!h]
\centering
\scriptsize
\setlength{\tabcolsep}{2.6pt}
\renewcommand{\arraystretch}{1.08}
\resizebox{\columnwidth}{!}{%
\begin{tabular}{|c|c|l|cccc|}
\hline
 & & Setting & ADE$\downarrow$ & FDE$\downarrow$ & MR$\downarrow$ & TSR$_{0.5}\uparrow$ \\
\hline

\multirow{14}{*}{\rotatebox{90}{Vehicle 1}}
& \multirow{7}{*}{\rotatebox{90}{valid}}
& Indiv &  1.123 & 2.098 & 45.78  & 37.88 \\
& & Fusion & \makecell{1.139 \\ \inc{\scriptsize{+0.016}}} & \makecell{2.139 \\ \inc{\scriptsize{+0.041}}} & \makecell{44.02 \\ \dec{\scriptsize{-1.76\%}}} & \makecell{39.09  \\ \incl{\scriptsize{+1.21\%}}} \\
& & Fusion + Delay  & \makecell{1.125 \\ \inc{\scriptsize{+0.002}}} & \makecell{2.099 \\ \inc{\scriptsize{+0.001}}} & \makecell{44.39 \\ \dec{\scriptsize{-1.39\%}}} & \makecell{39.06  \\ \incl{\scriptsize{+1.18\%}}} \\
& & Fusion + Delay + Packet Drop & \makecell{1.124 \\ \inc{\scriptsize{+0.001}}} & \makecell{2.096 \\ \dec{\scriptsize{-0.002}}} & \makecell{44.48 \\ \dec{\scriptsize{-1.3\%}}} & \makecell{39.02  \\ \incl{\scriptsize{+1.14\%}}} \\
\cline{2-7}
& \multirow{7}{*}{\rotatebox{90}{test}}
& Indiv & 1.524 & 3.045 & 54.44 & 24.88  \\
& & Fusion  & \makecell{1.517 \\ \dec{\scriptsize{-0.007}}} & \makecell{3.033 \\ \dec{\scriptsize{-0.012}}} & \makecell{52.77 \\ \dec{\scriptsize{-1.67\%}}}
 & \makecell{26.57 \\ \incl{\scriptsize{+1.69\%}}}
  \\
& & Fusion + Delay & \makecell{1.526 \\ \inc{\scriptsize{+0.002}}} & \makecell{3.043 \\ \dec{\scriptsize{-0.002}}} & \makecell{52.48 \\ \dec{\scriptsize{-1.96\%}}}
 & \makecell{26.49 \\ \incl{\scriptsize{+1.61\%}}}
  \\
& & Fusion + Delay + Packet Drop & \makecell{1.526 \\ \inc{\scriptsize{+0.002}}} & \makecell{3.043 \\ \dec{\scriptsize{-0.002}}} & \makecell{52.62 \\ \dec{\scriptsize{-1.82\%}}}
 & \makecell{26.47 \\ \incl{\scriptsize{+1.59\%}}} \\
\hline

\multirow{14}{*}{\rotatebox{90}{Vehicle 2}}
& \multirow{7}{*}{\rotatebox{90}{valid}}
& Indiv & 1.149 & 2.342 & 55.29 & 24.77  \\
& & Fusion  & \makecell{1.143 \\ \dec{\scriptsize{-0.006}}} & \makecell{2.330 \\ \dec{\scriptsize{-0.012}}} & \makecell{53.63 \\ \dec{\scriptsize{-1.66\%}}} & \makecell{26.46  \\ \incl{\scriptsize{+1.69\%}}} \\
& & Fusion + Delay   & \makecell{1.151 \\ \inc{\scriptsize{+0.002}}} & \makecell{2.340 \\ \dec{\scriptsize{-0.002}}} & \makecell{53.34 \\ \dec{\scriptsize{-1.95\%}}} & \makecell{26.37  \\ \incl{\scriptsize{+1.6\%}}} \\
& & Fusion + Delay + Packet Drop & \makecell{1.151 \\ \inc{\scriptsize{+0.002}}} & \makecell{2.340 \\ \dec{\scriptsize{-0.002}}} & \makecell{53.48 \\ \dec{\scriptsize{-1.81\%}}} & \makecell{26.35  \\ \incl{\scriptsize{+1.58\%}}} \\
\cline{2-7}

& \multirow{7}{*}{\rotatebox{90}{test}}
& Indiv & 1.462 & 2.717 & 45.11 & 37.78  \\
& & Fusion & \makecell{1.478 \\ \inc{\scriptsize{+0.016}}} & \makecell{2.757 \\ \inc{\scriptsize{+0.040}}} & \makecell{43.36 \\ \dec{\scriptsize{-1.75\%}}} & \makecell{39.00 \\ \incl{\scriptsize{+1.22\%}}} \\
& & Fusion + Delay  & \makecell{1.464 \\ \inc{\scriptsize{+0.002}}} & \makecell{2.718 \\ \inc{\scriptsize{+0.001}}} & \makecell{43.73 \\ \dec{\scriptsize{-1.38\%}}} & \makecell{38.95 \\ \incl{\scriptsize{+1.17\%}}} \\
& & Fusion + Delay + Packet Drop & \makecell{1.462 \\ \nog} & \makecell{2.715 \\ \dec{\scriptsize{-0.002}}} & \makecell{43.83 \\ \dec{\scriptsize{-1.28\%}}} & \makecell{38.91 \\ \incl{\scriptsize{+1.13\%}}} \\
\hline
\end{tabular}}
\caption{Trajectory prediction results on V2V4Real, comparing individual and collaborative prediction under communication delay and packet drop.}
\label{tab:v2v4real_time_delay}
\end{table}

For the valid split, Vehicle~1 shows an increase in missing objects, which leads to an improvement in displacement error compared with the fusion-only case. The TSR$_{0.5}$ rate shows only a mild drop of 0.07\% after introducing delay and packet drop. Vehicle~2 shows a different trend, with a decrease in missing objects when delay is introduced and a slight increase when packet drop is added, which leads to a decrease in displacement error. TSR$_{0.5}$ decreases steadily in this case. Similar trends are observed for the test split. Vehicle~1 shows a decrease in missing objects once delay is introduced, while Vehicle~2 follows a monotonic trend, with the largest drop in missing objects reaching 0.47\% and a moderate TSR$_{0.5}$ drop of 0.09\%. These results indicate that the proposed collaborative approach remains robust under communication delay and packet drop, with moderate attenuation of the gains achieved by fusion.

\begin{figure}[h!]
\centering
   \includegraphics[width=0.48\textwidth]{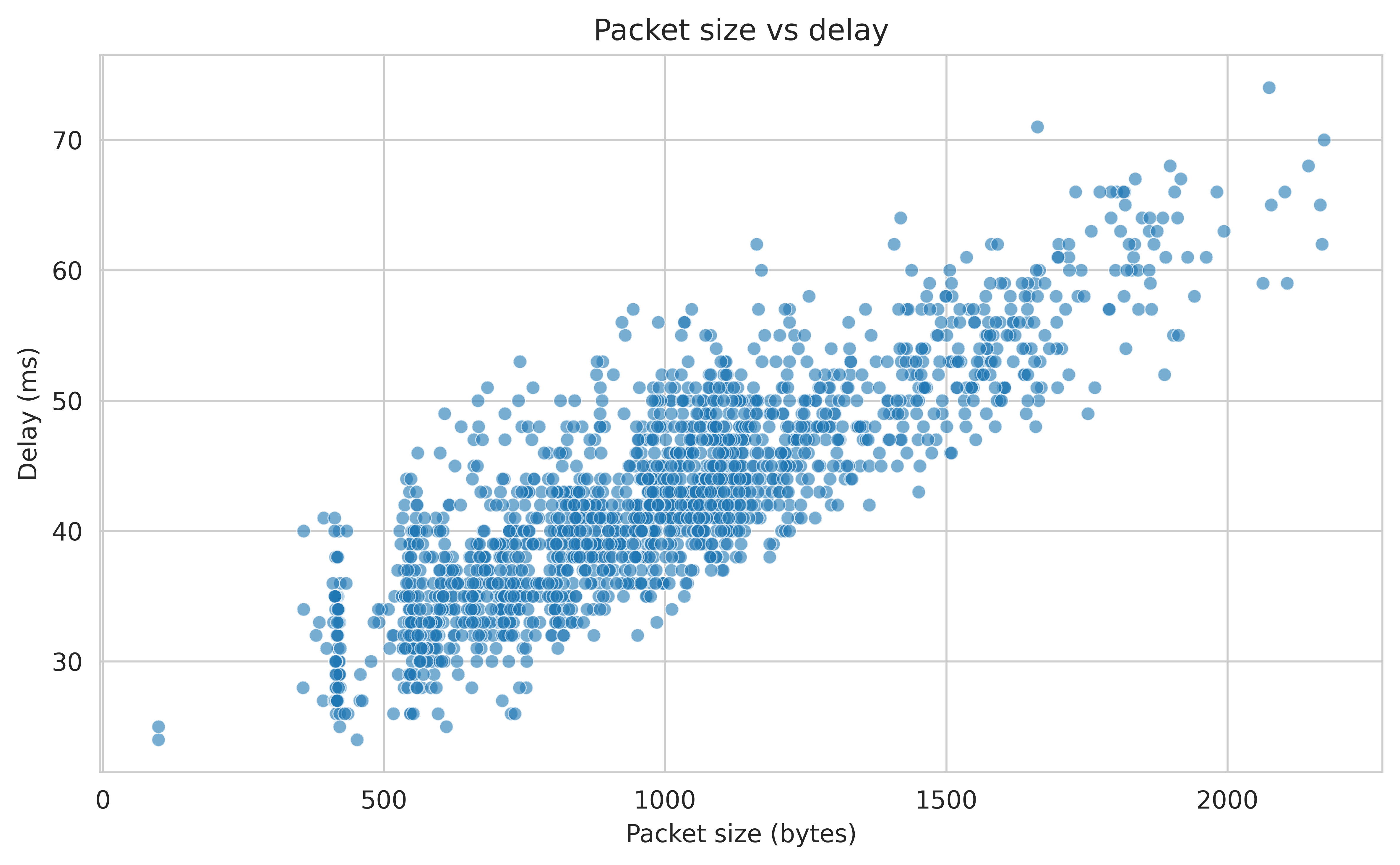}
\caption{Per-broadcast communication delay as a function of packet size.}
   \label{fig:delay_drop_packet_size}
\end{figure}

\begin{figure}[h!]
\centering
   \includegraphics[width=0.48\textwidth]{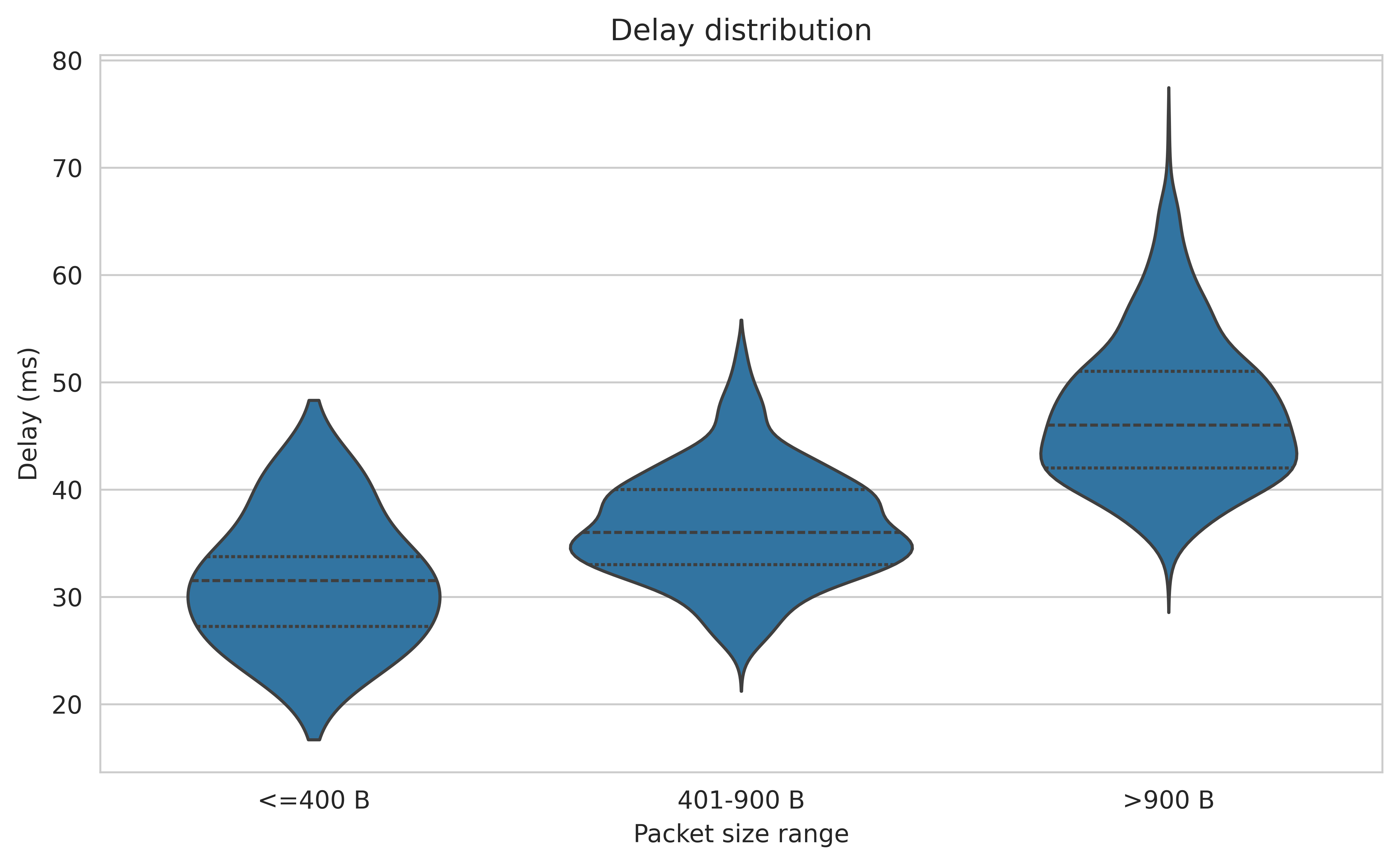}
\caption{Distribution of communication delay across three broadcast size categories.}
   \label{fig:delay_drop_distribution}
\end{figure}

Figure~\ref{fig:delay_drop_packet_size} confirms that the adopted communication model produces the intended delay behavior, where per-broadcast delay increases with packet size, while maintaining moderate variability due to the stochastic log-normal component. Figure~\ref{fig:delay_drop_distribution} further shows a consistent rightward shift of the delay distribution from smaller to larger packet-size categories, with only limited overlap between adjacent groups. 

%% file: sections/6_conclusion.tex
\section{Discussion and Conclusion} 
\label{sec:conclusion}

This paper proposed a collaborative framework for trajectory prediction. At its core is a collaboration layer based on a dynamic object-level map that associates local and  asynchronously received shared predictions, and defines the message structure for broadcasting. Built on top of this representation, the proposed late-fusion module combines local and shared trajectories to refine ego forecasts and reconstruct trajectories for occluded agents.

The conducted experiments show that the proposed framework consistently reduces the missing rate and improves $\mathrm{TSR}_{0.5}$ across all three datasets (OPV2V, DeepAccident, and V2V4Real). This trend remains stable under different predictor backbones and horizon configurations. In addition, the evaluation quantifies communication volume and examines the framework under realistic transmission effects, including latency and packet drop. On V2V4Real, the same overall trend is preserved when moving from the ray-casting occlusion setup to the full LiDAR--detector--tracker stack, indicating that the benefits of collaborative late fusion carry over to a more realistic perception pipeline.

Future work spans several directions. First, further investigation is needed into uncertainty estimation and calibration. If predictor-internal uncertainty can be decoupled from input-driven uncertainty, the fusion process could become more uncertainty-source aware. Both estimates could then be integrated into fusion to distinguish, on a per-agent basis, between shares arising from better observability and those arising from stronger motion forecasting. In addition, calibrated uncertainty could support an aggregation filter that prioritizes lower-uncertainty shares before fusion. Together, these extensions are expected to improve displacement errors.

Second, communication bandwidth can undergo further reduction. It can be formulated as an explicit optimization problem, selecting which agents, horizons, or message components to broadcast so that information gain is maximized while staying within the 5G NR-V2X budget. 

Finally, time feasibility invites further work on reducing the cost of Gaussian Process fitting. One potential direction is to use a gating mechanism to remove noisy shares and to precompute kernel parameters on filtered pools. This includes triggering expensive re-optimization only when the expected benefit justifies the latency.